\pdfoutput=1

\documentclass[11pt]{article}
\usepackage[dvipsnames]{xcolor}

\usepackage{ACL2023}

\usepackage{times}
\usepackage{latexsym}

\usepackage[T1]{fontenc}

\usepackage[utf8]{inputenc}

\usepackage{microtype}

\usepackage{inconsolata}

\usepackage{amsmath}
\usepackage{mathtools}
\usepackage{bbm}
\usepackage{multirow}
\usepackage{multicol}
\usepackage{makecell}
\usepackage{booktabs}
\usepackage{adjustbox}
\usepackage[linesnumbered,ruled,vlined]{algorithm2e}
\usepackage{comment}
\usepackage[shortlabels]{enumitem}
\usepackage{kotex}
\usepackage{amssymb}
\usepackage{listings}

\newcommand{\sub}[2]{
${\text{\textbf{#1}}\text{[}\textcolor{purple}{\textit{\textbf{#2}}}\text{]}}$
}
\newcommand\emp[1]{\textcolor{purple}{\textbf{#1}}}

\SetCommentSty{mycommfont}

\usepackage{cleveref}

\crefformat{section}{\S#2#1#3}
\crefformat{subsection}{\S#2#1#3}
\crefformat{subsubsection}{\S#2#1#3}

\usepackage{amsfonts}

\DeclareMathOperator{\EX}{\mathbb{E}}

\title{Robust Multi-bit Natural Language Watermarking \\through Invariant Features}

\author{KiYoon Yoo\textsuperscript{1}\hspace{.2cm} 
        Wonhyuk Ahn\textsuperscript{2}\hspace{.2cm}
        Jiho Jang\textsuperscript{1}\hspace{.2cm}
        Nojun Kwak\textsuperscript{1}\textsuperscript{*}\hspace{.2cm}
        \\
\textsuperscript{1}{Seoul National University}\hspace{.2cm}
\textsuperscript{2}{Webtoon AI} \\
\texttt{\{961230,geographic,nojunk\}@snu.ac.kr} \hspace{.2cm} \texttt{whahnize@gmail.com}
}

\begin{document}
\maketitle

\begin{abstract}
Recent years have witnessed a proliferation of valuable original natural language contents found in subscription-based media outlets, web novel platforms, and outputs of large language models. However, these contents are susceptible to illegal piracy and potential misuse without proper security measures. This calls for a secure watermarking system to guarantee copyright protection through leakage tracing or ownership identification. To effectively combat piracy and protect copyrights, a multi-bit watermarking framework should be able to embed adequate bits of information \textit{and} extract the watermarks in a robust manner despite possible corruption. In this work, we explore ways to advance both payload and robustness by following a well-known proposition from image watermarking and identify features in natural language that are invariant to minor corruption. Through a systematic analysis of the possible sources of errors, we further propose a corruption-resistant infill model. Our full method improves upon the previous work on robustness by +16.8\% point on average on four datasets, three corruption types, and two corruption ratios.\footnote{\noindent Department of Intelligence and Information, Graduate School of Convergence Science and Technology.\\https://github.com/bangawayoo/nlp-watermarking}
\end{abstract}

\section{Introduction}
Recent years have witnessed a proliferation of original and valuable natural language contents such as those found in subscription-based media outlets (e.g. Financial Times, Medium), web novel platforms (e.g. Wattpad, Radish) -- an industry that has shown rapid growth, especially in the East Asian market \citep{webnovel1, webnovel2} -- and texts written by human-like language models~\cite{chatgpt, vicuna2023, alpaca}. 
Without proper security measures, however, these contents are susceptible to illegal piracy and distribution, financially damaging the creators of the content and the market industry. 
In addition, the recent emergence of human-like language models like ChatGPT has raised concerns regarding the mass generation of disinformation~\cite{goldstein2023generative}.
This calls for a secure watermarking system to guarantee copyright protection or detect misuse of language models.

Digital watermarking is a technology that enables the embedding of information into multimedia (e.g. image, video, audio) in an unnoticeable way without degrading the original utility of the content. Through embedding information such as owner/purchaser ID, its application includes leakage tracing, ownership identification, meta-data binding, and tamper-proofing.
To effectively combat intentional evasion by the adversary or unintentional digital degradation, a watermarking framework should not only be able to embed adequate bits of information but also demonstrate robustness against potential corruption~\cite{tao2014robust, zhu2018hidden}. 
Watermarking in image and video contents has been extensively explored for pre-deep learning methods ~\citep{hsu1999hidden, wolfgang1999perceptual, wang2001image}. With the advent of deep neural networks, deep watermarking has emerged as a new paradigm that improves the three key aspects of watermarking: payload (i.e. the number of bits embedded), robustness (i.e. accuracy of the extracted message), and quality of the embedded media.

Natural language watermarking uses text as the carrier for the watermark by imperceptibly modifying semantics and/or syntactic features. As opposed to altering the visual appearances~\cite{rizzo2019fine}, this type of modification makes natural language watermarking resistant to piracy based on manual transcription. Previous research has focused on techniques such as lexical substitution with predefined rules and dictionaries or structural transformation~\citep{topkara2006natural, topkara2006hiding, atallah2001natural}.
Through utilizing neural networks, recent works have either replaced the predefined set of rules with learning-based methodology~\cite[AWT]{abdelnabi2021adversarial}, thereby removing heuristics or vastly improved the quality of lexical substitution~\cite[ContextLS]{yang2022tracing}. Despite the superiority over traditional methods, however, recent works are not without their limitations: AWT is prone to error during message extraction especially when a higher number of bits are embedded and occasionally generates deteriorated watermarked samples due to its entire reliance 
on a neural network; ContextLS has a fixed upper-bound on the payload and more importantly, does not consider extracting the bit message under corruption, which leads to low robustness. This work strives to advance both payload and robustness of natural language watermarking.

To build an effective robust watermarking system for natural language, we draw inspiration from a well-known proposition of a classical image watermarking work~\citep{cox1997secure}: That watermarks should \textit{"be placed explicitly in the perceptually most significant components"} of an image. If this is achieved, the adversary must corrupt the content's fundamental structure to destroy the watermark. 
This degrades the utility of the original content, rendering the purpose of pirating futile. 

However, embedding the watermark directly on the "perceptually most significant components" is only possible for images due to the inherent perceptual capacity of images. That is, modification in individual pixels is much more imperceptible than on individual words. Due to this, while we adhere to the gist of the proposition, we do not embed directly on the most significant component. Instead, we identify features that are semantically or syntactically fundamental components of the text and thus, invariant to minor modifications in texts. Then we use them as anchor points to pinpoint the position of watermarks. 
After formulating a general framework for robust natural watermarking, we empirically study the effectiveness of various potential invariant features derived from the semantic and syntactic components. Through step-by-step analysis of the possible sources of errors during watermark extraction, we further propose a corruption-resistant infill model that is trained explicitly to be robust on possible types of corruption.

Our experimental results encompassing four datasets of various writing styles demonstrate the robustness of (1) relying on invariant features for watermark embedding (2) using a robustly trained infill model. The absolute robustness improvement of our full method compared with the previous work is +16.8\% point on average on the four datasets, three corruption types, and two corruption ratios.

\begin{figure*}[t]
    \centering
    \includegraphics[width=1.\textwidth]{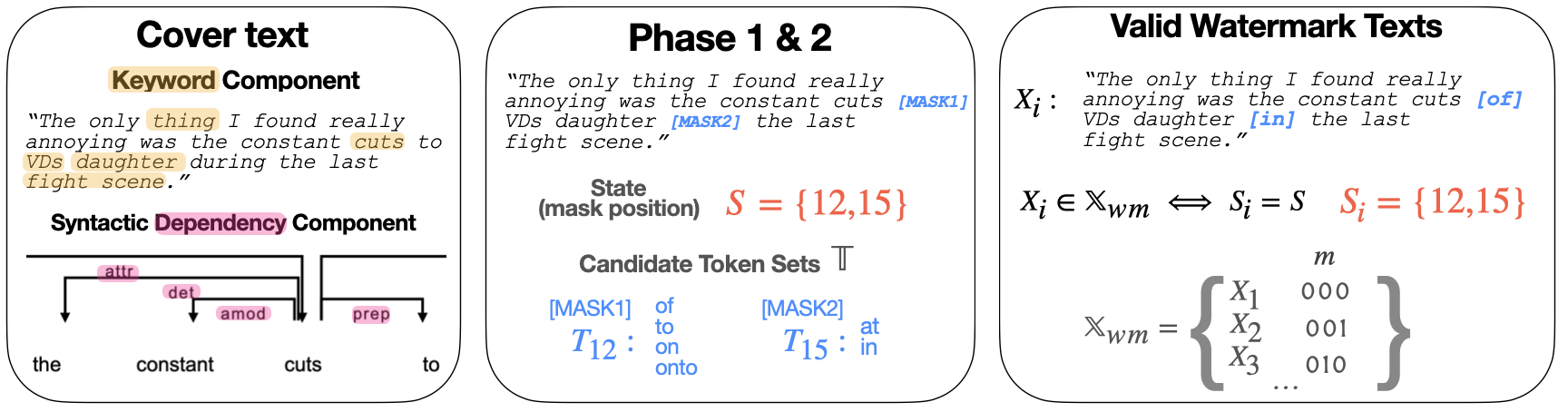}
    \caption{Leftmost shows an example of a cover text and its keyword and syntactic dependency components (only partially shown due to space constraint); Middle shows Phase 1 and Phase 2; Rightmost shows an example of a valid watermark sample.} 
    \label{fig1}
\end{figure*}

\section{Preliminaries}
\subsection{Problem Formulation of Watermarking }
In watermarking, the sender embeds a secret message $m$ into the cover text $X$ to attain the watermarked text $X_{\text{wm}}=\textsc{embed}(X,m)$. A cover text is the original document that is to be protected. A message, for instance, can be the ID of a purchaser or owner of the document represented in bit.
The receiver\footnote{Contrary to the separate terms (the sender and receiver) the two parties may be identical.} attempts to extract the embedded message  $\hat{m}=\textsc{extract}(\tilde{X}_{\text{wm}})$ from $\tilde{X}_{\text{wm}}=\textsc{corrupt}(X_{\text{wm}})$ which may be corrupted via intentional tampering by an adversary party as well as to natural degradation (e.g. typo) that may occur during distribution. We focus on blind watermarking, which has no access to the original cover text. The main objectives of the sender and the receiver are (1) to attain $X_{\text{wm}}$ that is semantically as similar as $X$ so as not to degrade the utility of the original content and (2) to devise the \textit{embed} and \textit{extract} functions such that the extracted message is accurate. 

\subsection{Corruptions on $X_{\text{wm}}$}
\label{subsec:corruption}
Conversely, the adversary attempts to interfere with the message extraction phase by corrupting the watermarked text, while maintaining the original utility of the text. For instance, an illegal pirating party will want to avoid the watermark being used to trace the leakage point while still wanting to preserve the text for illegal distribution. This constrains the adversary from corrupting the text too much both quantitatively and qualitatively. To this end, we borrow techniques from adversarial attack~\cite{jin2020bert, morris2020reevaluating} to alter the text and maintain its original semantics. 

We consider word insertion~\cite{li2021contextualized}, deletion~\cite{fengpathologies}, and substitution~\cite{garg2020bae} across 2.5\% to 5.0\% corruption ratios of the number of words in each sentence following~\citet{abdelnabi2021adversarial}. The number of words inserted/substituted/deleted is equal to $\textsc{round}(CR\times N)$ where $CR$ is the corruption ratio and $N$ is the number of words in the sentence. This ensures shorter sentences containing little to no room for corruption are not severely degraded. To additionally constrain the corrupted text from diverging from the original text, we use the pre-trained sentence transformer\footnote{https://www.sbert.net/} \textit{all-MiniLM-L6-v2}, which was trained on multiple datasets consisting of 1 billion pairs of sentences, to filter out corrupted texts that have cosine similarity less than 0.98 with the original text. 

\subsection{Infill Model}\label{subsec:infill}
Similar to ContextLS~\cite{yang2022tracing}, we use a pre-trained infill model to generate the candidates of watermarked sets. 
Given a masked sequence $X_{\backslash i}=\{x_1, \cdots, x_{i-1}, \text{MASK}, x_{i+1}, \cdots, x_{t}\}$, an infill language model can predict the appropriate words to fill in the mask(s). An infill model parameterized by $\theta$ outputs the probability distribution of $x_i$ over the vocabulary ($v$):  
\begin{equation}
\label{eq:infill_model}
    P(X_{\backslash i} | \theta) = p_i \in \mathbb{R}^{|v|}_+ .
\end{equation}
We denote the set of top-$k$ token candidates outputted by the infill model as 
\begin{equation}
\label{eq:infill_output}
    \{t_1^i, \cdots, t_k^i\} = \textsc{infill}(X_{\backslash i}; k).
\end{equation}

\section{Framework for Robust Natural Language Watermarking}
Our framework for natural language watermarking is composed of two phases. Phase 1 is obtaining state $S$ from the text $X$ (or $\tilde{X}_{\text{wm}}$) using some function $g_1$. $S$ can be considered as the feature abstracted from the text \textit{that contains sufficient information} to determine the embedding process. Phase 2 comprises function $g_2$ that takes $X$ and $S$ as inputs to generate the valid watermarked texts.
We rely on the mask infilling model to generate the watermarked texts, which makes $S$ the positions of the masks. The infill model generates the watermarked text $X_{\text{wm}}$ depending on the bit message. A general overview is shown in Figure \ref{fig1}.

\subsection{Phase 1: Mask Position Selection}
\label{subsec:Phase1}

For the watermarking system to be robust against corruption, $S$ should be chosen such that it depends on the properties of the text that are relatively invariant to corruption. That is, $S$ should be a function of the \textit{invariant features} of the text.
More concretely, an ideal \textit{invariant feature} is characterized by:
\begin{enumerate}
    \item A significant portion of the text has to be modified for it to be altered. 
    \vspace{-2mm}
    \item Thus, it is invariant to the corruptions that preserve the utility (e.g. semantics, nuance) of the original text.
\end{enumerate}
By construction, when $S$ is a function of an ideal invariant feature, this allows recovering the identical state $S$ for both $X$ and $\tilde{X}_{\text{wm}}$, which will enhance the robustness of the watermark. In essence, we are trying to find which words should be masked for the watermark to be robust.

Given a state function $g_1(\cdot)$, let $S=g_1(X)$, $\tilde{S}=g_1(\tilde{X}_{\text{wm}})$. Then, we define the \textbf{robustness of $g_1$} as follows:
\begin{equation}
\label{eq:state_robustness}
    \mathcal{R}_{g_1} \coloneqq \EX[\mathbbm{1}(S=\tilde{S})].
\end{equation}
Here, $\mathbbm{1}$ denotes the indicator function and $\EX$ is the expectation operation.

We sought to discover invariant features in the two easily attainable domains in natural language: semantic and syntactic components. An illustration of these components is shown in Figure \ref{fig1} Left.

\noindent\textbf{Keyword Component}
On the semantic level, we first pinpoint keywords that ought to be maintained for the utility of the original text to be maintained. Our intuition is that keywords are semantically fundamental parts of a sentence and thus, are maintained and invariant 
despite corruption. This includes proper nouns as they are often not replaceable with synonyms without changing the semantics (e.g. name of a movie, person, region), which can be extracted by an off-the-shelf Named Entity Recognition model. In addition, we use an unsupervised method called YAKE~\citep{campos2018yake} that outputs semantically essential words.  After extracting the keywords, we use them as anchors and can determine the position of the masks by a simple heuristic. For instance, the word adjacent to the keyword can be selected as the mask. 

\begin{table}[t]
\begin{adjustbox}{width=\columnwidth}
\begin{tabular}{ccccc}
    \toprule
    Robustness & \makecell{\small{Corr.}\\ \small{Types}} & \makecell{ContextLS\\\citep{yang2022tracing}} & Keyword & Syntactic  \\
    \hline
    \multirow{3}{*}{$\mathcal{R}_{g_1}$} 
                            & D & 0.656 & 0.944 & 0.921   \\
                            & I & 0.608 & 0.955 & 0.959   \\
                            & S & 0.646 & 0.974 & 0.949    \\
    \bottomrule
\end{tabular}
\end{adjustbox}
\caption{Robustness of $g_1$ ($ \mathcal{R}_{g_1}$) for ContextLS and Ours (Keyword, Syntactic) against three corruption types: Deletion (D), Insertion (I), and Substitution (S) under 5\% corruption rate on IMDB. See Appendix Table \ref{table:robustness_full} for full results.}
\label{tab:robustness_simple}
\vspace{-5mm}
\end{table}

\noindent\textbf{Syntactic Dependency Component} 
On the syntactic level, we construct a dependency parsing tree employing an off-the-shelf parser. A dependency parser describes the syntactic structure of a sentence by constructing a directed edge between a head word and its dependent word(s). Each dependent word is labeled as a specific type of dependency determined by its grammatical role. We hypothesize that the overall grammatical structure outputted by the parsing tree will be relatively robust to minor corruptions in the sentence. To select which type of dependency should be masked, we construct a predefined ordering to maintain the semantics of the watermarked sentences. The ordering is constructed by masking and substituting each type of dependency using an infill model and comparing its entailment score computed by an NLI model(e.g. RoBERTa-Large-NLI\footnote{https://huggingface.co/roberta-large-mnli}) on a separate held-out dataset as shown in Alg. \ref{alg1} (a more detailed procedure and the full list are provided in the Appendix \ref{appendix:nli-ordering}). Using the generated ordering, we mask each dependency until the target number of masks is reached. For both types of components (semantic \& syntactic), we ensure that keywords are not masked. 

So how well do the aforementioned components fare against corruption?
The results in Table \ref{tab:robustness_simple} bolster our hypothesis that keywords and syntactic components may indeed act as invariant features as both show considerably high robustness across three different types of corruption measured by the ratio of mask matching samples. As opposed to this, ContexLS~\citep{yang2022tracing}, which does not rely on any invariant features has a drastically lower $\mathcal{R}_{g_1}$. This signifies that a different word is masked out due to the corruption, which hampers the watermark extraction process.

\SetKwInput{KwInput}{Input}              
\SetKwInput{KwOutput}{Output}  
\maketitle
\begin{algorithm}[t]
\small
\DontPrintSemicolon
    \KwInput{Sentence $X$}
    \KwOutput{Sorted list $L$}

    \tcc{Find dependency of each word in $x\in X$ using Spacy}
    $x.\texttt{dep} \leftarrow \textsc{spacy}(X,x)$\;
    \tcc{Initiate dictionary of lists per dependency type}
    $D[x.\texttt{dep} ]:[~]$\;
    $N \leftarrow \texttt{len}(X)$\;
    \tcc{Loop through words and infill}
    \For{$i \gets 0$\texttt{ to } $N$}
        {
        $X' \leftarrow \texttt{INFILL}(X_{\backslash i})$\;
        $s \leftarrow \texttt{NLI}(X', X)$\;
        $D[x.\texttt{dep} ].\texttt{append}(s)$
        }

    \For{$\texttt{v}\in D\texttt{.values()}$}
        {
        $\texttt{v} \leftarrow \texttt{v}.\texttt{mean()}$\; 
        }
        
    $L=\texttt{sorted(}$\parbox[t]{0.8\linewidth}{
        $\texttt{[k for k,v in } D.\texttt{items()]}$,\\
        $\texttt{key=lambda x:x[1])}$ \;}
    \Return $L[::-1]$

\caption{Sorting syntactic dependency based on the NLI entailment score.}
\label{alg1}
\end{algorithm}

\subsection{Phase 2: Watermark Encoding}
\label{subsec:Phase2}

In Phase 2, a set of valid watermarked texts is generated by $g_2(X, S)$ to embed or extract the message. For ours, since the state is the set of mask positions, this comprises using an infill model to select top-$k$ words and alphabetically sort them to generate a valid set of watermarks.
Concretely, using the notations from \cref{subsec:infill}, $g_2(X, S)$ can be divided into the following steps: 
\begin{enumerate}[(1)]
    \item $\begin{aligned}[t]
    \mathcal{T}_i &=\{t_1^i, \cdots, t_k^i\}  
                  = \textsc{infill}(X_{\backslash i} ; k_1), 
                  \forall i \in S
    \end{aligned}$
    \item Filter $\mathcal{T}_i$ to remove any punctuation marks, subwords, stopwords. Update $\mathcal{T}_i$ by selecting top-$k_2$ ($\le k_1$) and sort them alphabetically.
    \item Form a cartesian product of the token sets $\mathbb{T}=  \mathcal{T}_{s_{1}} \times \cdots \times \mathcal{T}_{s_{j}}$ where $j=|S|$. Let $\mathbb{X}$ be the set of texts with the corresponding tokens substituted $(|\mathbb{X}|=|\mathbb{T}|)$.  
    \item Generate a \textit{valid} watermarked set $\mathbb{X}_{\textrm{wm}} = \{X_i \in \mathbb{X} | g_1(X_{wm})=g_1(X_i)\} \subseteq \mathbb{X}$ and assign a bit message for each element in the set $\mathbb{X}_{\textrm{wm}}$.
\end{enumerate}
In (4), generating a \textit{valid} set of watermarks means ensuring the message bit can be extracted without any error. This is done by keeping only those watermarked texts from $\mathbb{X}$ that have the same state as $X$ (Figure \ref{fig1} Middle and Right).
Under zero corruption (when $X_{\text wm}$=$\tilde{X}_{\text{wm}}$), Phase 2 will generate the same sets of watermarked texts if $S$ and $\tilde{S}$ are equivalent (i.e. $g_2(X, S)=g_2(\tilde{X}_{\text{wm}}, \tilde{S})$). Thus, our method is able to extract the watermark without any error when there is no corruption.

However, what happens when there \textit{is} corruption in the watermarked texts? Even if the exact state is recovered, the same set of watermarked texts may not be recovered as the infill model relies on local contexts to fill in the masks. Noting this in mind, we can also define the \textbf{robustness of $g_2$} as 
\begin{equation}
\label{eq:phase2_robustness}
    \mathcal{R}_{g_2} \coloneqq \EX[\mathbbm{1}(g_2(X, S)=g_2(\tilde{X}_{\text{wm}}, \tilde{S}))].
\end{equation}

\begin{figure}[t]
    \centering
    \includegraphics[width=0.48\textwidth]{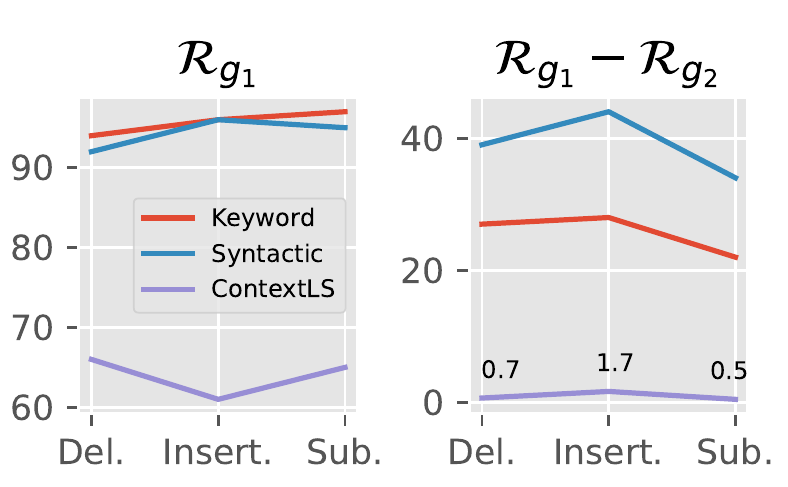}
    \caption{Robustness of $g_1$ and the difference between robustness of $g_1$ and $g_2$ under 5\% corruption rate on IMDB.} 
    \label{fig:robustness}
\end{figure}

Figure \ref{fig:robustness} Right shows $\mathcal{R}_{g_1}$ and the difference between $\mathcal{R}_{g_1}$ and $\mathcal{R}_{g_2}$. We observe that $\mathcal{R}_{g_2}$ is significantly lower than $\mathcal{R}_{g_1}$ for ours when we choose the infill model to be a vanilla pretrained language model such as BERT. While the type of invariant features does influence $\mathcal{R}_{g_2}$, our key takeaway is that $\mathcal{R}_{g_2}$ is substantially lower than $\mathcal{R}_{g_1}$ in all cases\footnote{Larger $\mathcal{R}_{g_2}$ does not necessarily imply a lower bit error rate as the extent of the discrepancy between $g_2(X, S)$ and $g_2(\tilde{X}_{\text{wm}}, \tilde{S})$ is not measured in the metric.}.

Interestingly, for ContextLS the gap between $\mathcal{R}_{g_1}$ and $\mathcal{R}_{g_2}$ is nearly zero, showing that Phase 1 is already a bottleneck for achieving robustness. The smaller gap can be explained by the use of smaller top-$k_2$(=2) and the incremental watermarking scheme, which incrementally increases the sequence to infill. This may reduce the possibility of a corrupted word influencing the infill model.

\begin{table}
\centering
\begin{adjustbox}{width=0.65\columnwidth}
\begin{tabular}{c|cc}
    \toprule
    Dataset & $\Delta\mathcal{R}_{g_1}$  & $\Delta\mathcal{R}_{g_2}$ \\
    D1 & .005\small{$\pm$.004} & .113\small{$\pm$.013}   \\
    D2 & .009\small{$\pm$.007} & .070\small{$\pm$.024}   \\
    D3 & .0\small{$\pm$.002} & .142\small{$\pm$.051}   \\
    D4 & .0\small{$\pm$.002} & .151\small{$\pm$.048}   \\
    \bottomrule
\end{tabular}
\end{adjustbox}
\caption{Effect of applying robust infill model on the robustness of Phase 1 and 2 (With - Without) averaged over the three corruption types up to three decimal points. The four datasets (D1 - D4) are IMDB, Wikitext-2, Dracula, and Wuthering Heights, respectively. Further details about the datasets are in \cref{sec:exp}.}
\label{tab:RI-effect}
\end{table}

\subsection{Robust Infill Model}
To overhaul the fragility of Phase 2, we build an infill model robust to possible corruptions by finetuning $\theta$ to output a consistent word distribution when given $X_{\backslash i}$ and $\tilde{X}_{\backslash i}$, a corrupted version of $X_{\backslash i}$. This can be achieved by minimizing the divergence of the two distributions $p_i$ and $\tilde{p}_i$ where $\tilde{p}_i$ refers to the word distribution of the corrupted sequence, $\tilde{X}_{\backslash i}$.  Instead of using the original word distribution as the target distribution, which is densely populated over $>$ 30,000 tokens (for BERT-base), we form a sparse target distribution over the top-$k_1$ tokens by zeroing out the rest of the tokens and normalizing over the $k_1$ tokens. This is because only the top-$k_{1}$ tokens are used in our watermarking frame (see \cref{subsec:Phase2}). 

In addition, to improve the training dynamics, we follow the masking strategy proposed in \cref{subsec:Phase1} to choose the words to masks, instead of following the random masking strategy used in the original pretraining phase. This aligns distributions of the masked words at train time and test time, which leads to a better performance (robustness) given the same compute time. As opposed to this, since the original masking strategy randomly selects a certain proportion of words to mask out, this will provide a weaker signal for the infill model to follow.

We use the Kullback–Leibler (KL) divergence as our metric. More specifically, we use the `reverse KL' as our loss term in which the predicted distribution (as opposed to the target distribution) is used to weigh the difference of the log distribution as done in Variational Bayes~\cite{kingmaauto}. This aids the model from outputting a "zero-forcing" predicted distribution.
The consistency loss between the two distributions is defined by 
\begin{align}
    \mathcal{L}_{con} &= \sum_{i \in S} \text{KL}(\tilde{p_i}|p_i), \\ 
    \text{where}~~~\tilde{p_i} &= P(\tilde{X}_{\backslash i} | \theta), \\
    p_i &= P(X_{\backslash i} | \textsc{freeze}(\theta))  
\end{align}
for all $i$ of the masked tokens. The graph outputting $p$ is detached to train a model to output a consistent output when given a corrupted input. As we expected, using the robust infill model to the Syntactic component leads to a noticeable improvement in $\mathcal{R}_{g_2}$, while that of $\mathcal{R}_{g_1}$ is negligible (Table \ref{tab:RI-effect}). 

 The corrupted inputs are generated following the same strategy in \cref{subsec:corruption} using a separate train dataset. We ablate our design choices in \cref{subsec:ablation}. 

 To summarize, the proposed framework
\begin{enumerate}
    \vspace{-1mm}
    \item allows the embedding and extraction of watermarks faultlessly when there is no corruption.
    \vspace{-8mm}
    \item can incorporate invariant features for watermark embedding, achieving robustness in the presence of corruption.
    \vspace{-3mm}
    \item further enhance robustness in Phase 2 by utilizing a robust infill model.
\end{enumerate}
\vspace{-3mm}

\begin{table}[t]
\begin{adjustbox}{width=\columnwidth}
\begin{tabular}{cc|c|ccc}
    \toprule
    \multicolumn{2}{c}{} & \multicolumn{4}{c}{\textbf{IMDB}} \\
    \multicolumn{2}{c}{} & \multicolumn{4}{c}{Methods} \\ 
    \cline{3-6}
     \multicolumn{2}{c}{Metrics} & ContextLS & Keyword & Syntactic & +RI    \\
    \hline \hline
    BPW ($\uparrow$) & & 0.100 & 0.116 & 0.125 & \textbf{0.144}   \\
    \hline
    \multirow{3}{*}{\makecell{BER($\downarrow$)\\@CR=0.025}} 
                & D  &  0.219  & 0.127 &  0.100 & \textbf{0.074} \\
                & I  &  0.303 &  0.153 &  0.153 & \textbf{0.106}\\
                & S  &  0.273 &  0.142 &  0.133 & \textbf{0.110}  \\
    \hline 
    \multirow{3}{*}{\makecell{BER($\downarrow$)\\@CR=0.05}} 
                & D  & 0.392 & 0.252  &  0.277  & \textbf{0.200} \\
                & I  & 0.355 & 0.201  &  0.242  & \textbf{0.163}  \\
                & S  & 0.343 & 0.218  &  0.220  & \textbf{0.177} \\
    \hline
    \end{tabular}
    \end{adjustbox} 
\begin{adjustbox}{width=\columnwidth}
\begin{tabular}{lc|cc|ccc}
    \multicolumn{2}{c}{} & \multicolumn{4}{c}{\textbf{Wikitext-2}} \\
    \multicolumn{2}{c}{} & \multicolumn{5}{c}{Methods} \\ 
    \cline{3-7}
    \multicolumn{2}{c}{Metrics} & AWT &  ContextLS & Keyword & Syntactic & +RI    \\
    \hline \hline
    BPW ($\uparrow$) &          &0.100& 0.083 & 0.092 & 0.090 & \textbf{0.136}\\
    \hline
    \makecell{BER($\downarrow$)@CR=0} & & 0.264 & 0.0  & 0.  & 0. & 0. \\
    \hline 
    \multirow{3}{*}{\makecell{BER($\downarrow$)\\@CR=0.025}} 
                & D  & 0.273 & 0.224 & 0.202  &  0.162  & \textbf{0.136} \\
                & I  & 0.272 & 0.289 & 0.222  &  0.216  & \textbf{0.205} \\
                & S  & 0.279 & 0.266 & 0.176  &  0.155  & \textbf{0.157} \\
    \hline 
    \multirow{3}{*}{\makecell{BER($\downarrow$)\\@CR=0.05}} 
                & D  & 0.284 & 0.410 & 0.326  &  0.321  & \textbf{0.282} \\
                & I  & 0.272 & 0.338 & 0.246  &  0.235  & \textbf{0.201} \\
                & S  & 0.289 & 0.342 & 0.256  &  0.228  & \textbf{0.201} \\
    \hline \hline
    \multicolumn{2}{c}{} & \multicolumn{4}{c}{\textbf{Dracula}} \\
    BPW ($\uparrow$)& & 0.100 & 0.089 &  0.126 & 0.117 & \textbf{0.146}    \\
    \hline
    \makecell{BER($\downarrow$)@CR=0}  & & 0.111 & 0.  & 0. & 0. & 0. \\
    \hline 
    \multirow{3}{*}{\makecell{BER($\downarrow$)\\@CR=0.025}} 
                & D  & 0.236 & 0.201 &  0.116 &  0.076 & \textbf{0.030} \\
                & I  & 0.218 & 0.299 &  0.181 &  0.133 & \textbf{0.063}  \\
                & S  & 0.231 & 0.272 &  0.140 &  0.130 & \textbf{0.081}   \\
    \hline 
    \multirow{3}{*}{\makecell{BER($\downarrow$)\\@CR=0.05}} 
                & D  & 0.286 & 0.373 & 0.255 &  0.248 & \textbf{0.177} \\
                & I  & 0.264 & 0.375 & 0.228 &  0.279 & \textbf{0.155}   \\
                & S  & 0.281 & 0.337 & 0.207 &  0.229 & \textbf{0.164} \\  
    \hline \hline
    \multicolumn{2}{c}{} & \multicolumn{4}{c}{\textbf{Wuthering Heights}} \\
    BPW ($\uparrow$)&  & 0.100 & 0.076 &  0.088 &  0.097 &  \textbf{0.114}   \\
    \hline
    \makecell{BER($\downarrow$)@CR=0}  & & 0.100 & 0.  & 0. & 0. & 0. \\
    \hline
    \multirow{3}{*}{\makecell{BER($\downarrow$)\\@CR=0.025}} 
                & D  & 0.224 & 0.194 &  0.102  &  0.088 & \textbf{0.063} \\
                & I  & 0.212 & 0.284 &  0.144  &  0.132 & \textbf{0.068}   \\
                & S  & 0.224 & 0.271 &  0.161  &  0.143 & \textbf{0.096}  \\
    \hline 
    \multirow{3}{*}{\makecell{BER($\downarrow$)\\@CR=0.05}} 
                & D  & 0.283 & 0.379 &  0.253 &  0.240  & \textbf{0.169} \\
                & I  & 0.258 & 0.363 &  0.224 &  0.268  & \textbf{0.133}   \\
                & S  & 0.276 & 0.363 &  0.231 &  0.245  & \textbf{0.161}   \\
    \bottomrule
\end{tabular}
\end{adjustbox}
\caption{Comparison of payload and robustness on four datasets. +RI denotes adding the robust infill model to our Syntactic component. \textbf{Top-1} numbers are shown in bold.}
\label{tab:main_robustness}
\end{table}

\section{Experiment}
\label{sec:exp}

\textbf{Dataset}
To evaluate the effectiveness of the proposed method, we use four datasets with various styles. IMDB~\citep{maas-EtAl:2011:ACL-HLT2011} is a movie reviews dataset, making it more colloquial. WikiText-2~\cite{merity2016pointer}, consisting of articles from Wikipedia, has a more informative style. We also experiment with two novels, Dracula and Wuthering Heights (WH), which have a distinct style compared to modern English and are available on Project Gutenberg~\citep{dracula,wh}. 

\noindent\textbf{Metrics}
For payload, we compute bits per word (BPW). For robustness, we compute the bit error (BER) of the extracted message. We also measure the quality of the watermarked text by comparing it with the original cover text. Following \citet{yang2022tracing, abdelnabi2021adversarial}, we compute the entailment score (ES) using an NLI model (RoBERTa-Large-NLI) and semantic similarity (SS) by comparing the cosine similarity of the representations outputted by a pre-trained sentence transformer (stsb-RoBERTa-base-v2). We also conduct a human evaluation study to assess semantic quality.

\noindent\textbf{Implementation Details}
For ours and ContextLS~\citep{yang2022tracing}, both of which operate on individual sentences, we use the smallest off-the-shelf model (\textit{en-core-web-sm)} from Spacy~\citep{spacy2} to split the sentences. The same Spacy model is also used for NER (named entity recognizer) and building the dependency parser for ours. Both methods use BERT-base as the infill model and select top-32 ($k_1$) tokens. We set our payload to a similar degree with the compared method(s) by controlling the number of masks per sentence ($|S|$) and the top-$k_2$ tokens (\cref{subsec:Phase2}); these configurations for each dataset are shown in Appendix Table \ref{tab:configuration}. We watermark the first 5,000 sentences for each dataset and use TextAttack~\cite{morris2020textattack} to create corrupted samples. For robust infilling, we finetune BERT for 100 epochs on the individual datasets. For more details, refer to the Appendix. 

\noindent\textbf{Compared Methods}
We compare our method with deep learning-based methods~\cite[AWT]{abdelnabi2021adversarial}\cite[ContextLS]{yang2022tracing} for our experiments as pre-deep learning methods~\citep{topkara2006hiding, Hao2018} that are entirely rule-based have low payload and/or low semantic quality (later shown in Table \ref{tab:semantic}). More details about the compared methods are in \cref{sec:RW}.

\subsection{Main Experiments}
\vspace{-2mm}
Table \ref{tab:main_robustness} shows the watermarking results on all four datasets. Some challenges we faced during training AWT and our approach to overcoming this are detailed in Appendix \ref{appendix:awt}. Since the loss did not converge on IDMB for AWT as detailed in \cref{appendix:awt-imdb}, we omit the results for this. 

We test the robustness of each method on corruption ratios (CR) of 2.5\% and 5\%. For ours, we apply robust infilling for the Syntactic Dependency Component, which is indicated in the final column by +RI. AWT suffers less from a larger corruption rate and sometimes outperforms our methods without RI. 
However, the BER at zero corruption rate is non-negligible, which is crucial for a reliable watermarking system. In addition, we observe qualitatively that AWT often repeats words or replaces pronouns on the watermarked sets, which seems to provide signals for extracting the message -- this may provide a distinct signal for message extraction at the cost of severe quality degradation. Some examples are shown in Appendix \ref{appendix:examples} and Tab. \ref{tab:example2}-\ref{tab:example4}.

Our final model largely outperforms ContextLS in all the datasets and corruption rates. Additionally, both semantic and syntactic components are substantially more robust than ContextLS even without robust infilling in all the datasets. The absolute improvements in BER by using Syntactic component across corruption types with respect to ContextLS under CR=2.5\% are 13.6\%, 8.2\%, 14.4\%, and 12.9\% points for the four datasets respectively when using the Syntactic component; For CR=5\%, they are 10.0\%, 10.2\%, 11.0\%, and 11.7\% points. 

\begin{table}[t]
\centering
\begin{adjustbox}{width=0.5\textwidth}
\begin{tabular}{l cccccc}
    \toprule
      & & [1] & [2] & 
      \makecell{AWT} &
      \makecell{ContextLS} & 
      Ours  \\
    \hline
    \multirow{2}{*}{IMDB}
        & ES & 0.843 & 0.867 & 0.958 & 0.985 & 0.975  \\
        & SS & 0.916 & 0.943 & 0.973 & 0.982 & 0.981  \\ 
    \hline
    \multirow{2}{*}{Wikitext-2}
        & ES & 0.888 & 0.907 & 0.935 & 0.986 & 0.966  \\
        & SS & 0.941 & 0.945 & 0.991 & 0.989  & 0.993  \\ 
    \hline
    \multirow{2}{*}{Dracula}
        & ES & 0.869 &  0.915 & 0.869 & 0.985 & 0.963    \\
        & SS & 0.910 &  0.889 & 0.855 & 0.986 & 0.971 \\ 
    \hline
    \multirow{2}{*}{WH}
        & ES & 0.882 &  0.893 & 0.947 & 0.984 & 0.964    \\
        & SS & 0.929 &  0.934 & 0.968 & 0.989 & 0.975 \\ 
    \bottomrule
\end{tabular}
\end{adjustbox}
\caption{[1]: \citet{topkara2006hiding}, [2]: \citet{Hao2018}.
Semantic scores (ES: entailment score, SS: semantic similarity) of the watermarked sets in relation to the original cover text. All numbers except ours are from \citet{yang2022tracing}}
\vspace{-2mm}
\label{tab:semantic}
\end{table}

\begin{table}[t]
\centering
\begin{adjustbox}{width=0.4\textwidth}
\begin{tabular}{l cccc}
    \toprule
    Metrics & AWT & ContextLS & Ours  \\
    \hline
    Fluency$\Delta(\downarrow)$ & 1.32\small{$\pm$0.7} & 0.25\small{$\pm$0.4} & 0.26\small{$\pm$0.4} \\
    SS$(\uparrow)$ & 2.97\small{$\pm$0.8} & 4.22\small{$\pm$0.5} & 3.90\small{$\pm$0.8}\\ 
    \bottomrule
\end{tabular}
\end{adjustbox}
\caption{Human evaluation results on Likert scale (20 samples and 5 annotators).}
\vspace{-5mm}
\label{tab:semantic-human}
\end{table}

\subsection{Semantic Scores of Watermark}
Table \ref{tab:semantic} shows the results for semantic metrics. While our method falls behind ContextLS, we achieve better semantic scores than all the other methods while achieving robustness. ContextLS is able to maintain a high semantic similarity by explicitly using an NLI model to filter out candidate tokens. However, the accuracy of the extracted message severely deteriorates in the presence of corruption as shown in the previous section.  Using ordered dependencies sorted by the entailment score significantly increases the semantic metrics than using a randomly ordered one, denoted by "--NLI Ordering". The results are in Appendix Table \ref{tab:semantic_ours}.

We also conduct human evaluation comparing the fluency of the watermarked text and cover text (Fluency$\Delta$) and how much semantics is maintained (Semantic Similarity; SS) compared to the original cover text in Tab. \ref{tab:semantic-human}. The details of the experiment are in \cref{appendix:human-eval}.
This is aligned with our findings in automatic metrics, but shows a distinct gap between ours and AWT. Notably, the levels of fluency change of ours and ContextLS compared to the original cover text are nearly the same.

\section{Discussion}
\begin{table}[t]
\centering
\begin{adjustbox}{width=0.85\columnwidth}
    \begin{tabular}{cc|ccc}
    \toprule
         & top-$k_2$ & 2 & 3 & 4 \\
        \hline 
        \multirow{2}{*}{BPW}
           & \small ContextLS & 0.100 & 0.033 & 0.021    \\
           & Ours & 0.100 & 0.161 & 0.211 \\
        \hline
        \multirow{2}{*}{\small Forward Pass}
           & \small ContextLS & 1994 & 2386 &  2801   \\
           & Ours & 94 & 94 & 94 \\
        \hline
    \end{tabular}
    \end{adjustbox}
    \caption{The effect of top-$k_2$ on payload, \# of forward pass to the infill model, and wall clock time for ContextLS and ours on IMDB. We fix our keyword ratio to 0.11. }
\label{tab:other_comparison}
\vspace{-5mm}
\end{table}

\vspace{-2mm}
\subsection{Comparison with ContextLS}
\label{subsec:discussion_with_cls}
Some design choices we differ from ContextLS is top-$k_2 > 2$ which determines the number of candidate tokens per mask. We can increase the payload depending on the requirement by choosing a higher $k_2$. However, for ContextLS increasing $k_2$ counter-intuitively leads to a \textit{lower} payload. This is because ContextLS determines the valid watermark sets (those that can extract the message without error) with much stronger constraints (for details see Eq. 5,6,7 of \citet{yang2022tracing}). This also requires an exhaustive search over the whole sentence with an incrementally increasing window, which leads to a much longer embedding / extraction time due to the multiple forward passes of the neural network. For instance, the wall clock time of embedding in 1000 sentences on IMDB is more than 20 times on ContextLS (81 vs. 4 minutes). More results are summarized in Table \ref{tab:other_comparison}. Results for applying our robust infill model to ContextLS are in Appendix \ref{appendix:nli-ordering}.

\begin{table}[t]
\begin{tabular}{p{0.97\linewidth}}
    \hrule 
    \centering \textbf{Coordination} \tabularnewline
    \small{Sci-fi movies/TV are usually underfunded, under-appreciated \sub{and}{nor} misunderstood. (ES=0.996, SS=0.989)}\\[5pt]
   \small I thought the main villains were pretty well done \sub{and}{but} fairly well acted. (ES=0.994, SS=0.994)\\
    \hrule
    \centering \textbf{Named Entity} \tabularnewline
    \small The only reason this movie is not given a 1 (awful) vote is that the acting of both \sub{Ida}{Ada} Lupino and \sub{Robert}{Rob} Ryan is superb. (ES=0.993, SS=0.961)\\
    \small I have not seen any other movies from the "\sub{Crime}{Criminal} Doctor" series, so I can’t make any comparisons. (ES=0.994, SS=0.990)
    \hrule 
\end{tabular}
\vspace{-3mm}
\caption{Entailment score between the cover text and the watermarked text. The \sub{original}{watermarked} words are shown.}
\vspace{-3mm}
\label{tab:failure_example}
\end{table}

\vspace{-2mm}
\subsection{Pitfalls of Automatic Semantic Metrics}
\label{subsec:pitfall}
Although the automatic semantic metrics do provide a meaningful signal that aids in maintaining the original semantics, they do not show the full picture. First, the scores do not accurately reflect the change in semantics when substituting for the coordination dependency (e.g. and, or, nor, but, yet). As shown in Table \ref{tab:failure_example}, both the entailment score and semantic similarity score overlook some semantic changes that are easily perceptible by humans. This is also reflected in the sorted dependency list we constructed in \cref{subsec:Phase1} - the average NLI score after infilling a coordination dependency is 0.974, which is ranked second. An easy fix can be made by placing the coordination dependency at the last rank or simply discarding it. We show in Appendix Table \ref{tab:dep-wo-cc-full} that this also provides a comparable BPW and robustness. 

Another pathology of the NLI model we observed was when a named entity such as a person or a region is masked out. Table \ref{tab:failure_example} shows an example in ContextLS and how ES is abnormally high. Such watermarks may significantly hurt the utility of novels if the name of a character is modified. This problem is circumvented in ours by disregarding named entities (detected using NER) as possible mask candidates.

\begin{table}[t]
\begin{adjustbox}{width=\columnwidth}
\begin{tabular}{c cccc}
    \toprule
      &  & Ran. Mask (FKL)& Ran. Mask (RKL) & Ours  \\
    \hline
    BPW($\uparrow$) & & 0.121 & 0.129 & 0.144 \\
    \hline
    \multirow{3}{*}{\makecell{BER($\downarrow$)\\@CR=0.025}} 
    & D & 0.106 & 0.101 & 0.074  \\
    & I & 0.141 & 0.139 & 0.106  \\
    & S & 0.138 & 0.137 & 0.110  \\
    \bottomrule
\end{tabular}
\end{adjustbox}
\caption{Ablation of masking design choices (FKL: Forward KL, RKL: Reverse KL). Ours is the final version used in the main experiments (our masking strategy + RKL).}
\label{tab:ablation}
\end{table}

\subsection{Ablations and Other Results}
\label{subsec:ablation}
\textbf{Ablations} In this section, we ablate some of the design choices. First, we compare the design choices of our masking strategies (random vs. ours) and loss terms (Forward KL and Reverse KL) in Table \ref{tab:ablation}. Our masking strategy improves both BPW and robustness compared to randomly masking out words. 
Though preliminary experiments showed RKL is more effective for higher payload and robustness, further experiments showed the types of KL do not significantly affect the final robustness when we use our masking strategy. We further present the results under character-based corruption and compare robustness against different corruption types in Appendix \ref{appendix:nli-ordering}. 

\noindent\textbf{Stress Testing Syntactic Component}
We experiment with how our proposed Syntactic component fares in a stronger corruption rate. The results are shown in Appendix Fig. \ref{fig:high-cr}. While the robustness is still over 0.9 for both insertion and substitution at CR=0.1, the robustness rapidly drops against deletion. This shows that our syntactic component is most fragile against deletion.

\section{Related Works}
\label{sec:RW}
Natural language watermarking embeds information via manipulation of semantics or syntactic features rather than altering the visual appearance of words, lines, and documents~\citep{rizzo2019fine}. This makes natural language watermarking robust to re-formatting of the file or manual transcription of the text \citep{topkara2005natural}. Early works in natural language watermarking have relied on synonym substitution~\citep{topkara2006hiding}, restructuring of syntactic structures~\citep{atallah2001natural}, or paraphrasing~\citep{atallah2003natural}. The reliance on a predefined set of rules often leads to a low bit capacity and the lack of contextual consideration during the embedding process may result in a degraded utility of the watermarked text that sounds unnatural or strange. 

With the advent of neural networks, some works have done away with the reliance on pre-defined sets of rules as done in previous works. Adversarial Watermarking Transformer \citep[AWT]{abdelnabi2021adversarial} propose an encode-decoder transformer architecture that learns to extract the message from the decoded watermarked text. To maintain the quality of the watermarked text, they use signals from sentence transformers and language models. However, due to entirely relying upon a neural network for message embedding and extraction, the extracted message is prone to error even without corruption, especially when the payload is high and has a noticeable artifact such as repeated tokens in some of the samples.
\citet{yang2022tracing} takes an algorithmic approach for embedding and extraction of messages, making it errorless. Additionally, using a neural infill model along with an NLI model has shown better quality in lexical substitution than more traditional approaches (e.g. WordNet). However, robustness under corruption is not considered. 

\noindent \textbf{Image Watermarking} Explicitly considering corruption for robustness and using different domains of the multimedia are all highly relevant to blind image watermarking, which has been extensively explored~\citep{mun2019finding, zhu2018hidden, zhong2020automated, luo2020distortion}. Like our robust infill training, \citeauthor{zhu2018hidden, luo2020distortion} explicitly consider possible image corruptions to improve robustness. 
Meanwhile, transforming the pixel domain to various frequency domains using transform methods such as Discrete Cosine Transform has shown to be both effective and more robust~\citep{potdar2005survey}. The use of keywords and dependencies to determine the embedding position in our work can be similarly considered as transforming the raw text into semantic and syntactic domains, respectively. 

\noindent \textbf{Other Lines of Work} Steganography is a similar line of work concealing secret data into a cover media focusing on covertness rather than robustness. Various methods have been studied in the natural language domain~\citep{tina2017generating, yang2018rnn, ziegler2019neural, yang2020vae, ueoka2021frustratingly}. This line of works differs from watermarking in that the cover text may be arbitrarily generated to conceal the secret message, which eases the constraint of maintaining the original semantics. 

Recently, \citet{he2022protecting} proposed to watermark outputs of language models to prevent model stealing and extraction. While the main objective of these works~(\citealp{he2022protecting, hecater}) differs from ours, the methodologies can be adapted to watermark text directly. However, these are only limited to zero-bit watermarking (e.g. whether the text is from a language model or not), while ours allow embedding of any multi-bit information. Similarly, \citet{kirchenbauer2023watermark} propose to watermark outputs of language models at decoding time in a zero-bit manner to distinguish machine-generated texts from human-written text. 

\section{Conclusion}
\vspace{-2mm}
We propose using invariant features of natural language to embed robust watermarks to corruptions. We empirically validate two potential components easily discoverable by off-the-shelf models. The proposed method outperforms recent neural network-based watermarking in robustness and payload while having a comparable semantic quality. We do not claim that the invariant features studied in this work are the optimal approach. Instead, we pave the way for future works to explore other effective domains and solutions following the framework.

\newpage
\section*{Limitations}
Despite its robustness, our method has subpar results on the automatic semantic metrics compared to the most recent work. This may be a natural consequence of the perceptibility vs. robustness trade-off~\cite{tao2014robust, de2002invisibility}: a stronger watermark tends to interfere with the original content. Nonetheless, by using some technical tricks (e.g. neural infill model, NLI-sorted ordering) our method is able to be superior to all the other methods including two traditional ones and a neural network-based method. 

Techniques from adversarial attack were employed to simulate possible corruptions in our work. However, these automatic attacks does not always lead to imperceptible modifications of the original texts~\citep{morris2020reevaluating}. Thus, the corruptions used in our work may be a rough estimate of what true adversaries might do to evade watermarking. In addition, our method is not tested against paraphrasing, which may substantially change the syntactic component of the text. One realistic reason that deterred us from experimenting on paraphrase-based attacks was their lack of controllability compared to other attacks that have fine-grained control over the number of corrupted words. Likewise, for text resources like novels that value subtle nuances, the aforementioned property may discourage the adversary from using it to destroy watermarking.

\section*{Acknowledgements}
This work was supported by Korean Government through the IITP grants 2022-0-00320, 2021-0-01343, NRF grant 2021R1A2C3006659 and by Webtoon AI at NAVER WEBTOON in 2022.

\bibliography{custom}

\begin{thebibliography}{47}
\expandafter\ifx\csname natexlab\endcsname\relax\def\natexlab#1{#1}\fi

\bibitem[{Abdelnabi and Fritz(2021)}]{abdelnabi2021adversarial}
Sahar Abdelnabi and Mario Fritz. 2021.
\newblock Adversarial watermarking transformer: Towards tracing text provenance
  with data hiding.
\newblock In \emph{2021 IEEE Symposium on Security and Privacy (SP)}, pages
  121--140. IEEE.

\bibitem[{Atallah et~al.(2001)Atallah, Raskin, Crogan, Hempelmann, Kerschbaum,
  Mohamed, and Naik}]{atallah2001natural}
Mikhail~J Atallah, Victor Raskin, Michael Crogan, Christian Hempelmann, Florian
  Kerschbaum, Dina Mohamed, and Sanket Naik. 2001.
\newblock Natural language watermarking: Design, analysis, and a
  proof-of-concept implementation.
\newblock In \emph{International Workshop on Information Hiding}, pages
  185--200. Springer.

\bibitem[{Atallah et~al.(2003)Atallah, Raskin, Hempelmann, Karahan, Sion,
  Topkara, and Triezenberg}]{atallah2003natural}
Mikhail~J Atallah, Victor Raskin, Christian~F Hempelmann, Mercan Karahan, Radu
  Sion, Umut Topkara, and Katrina~E Triezenberg. 2003.
\newblock Natural language watermarking and tamperproofing.
\newblock In \emph{International workshop on information hiding}, pages
  196--212. Springer.

\bibitem[{Bram(1897)}]{dracula}
Stoker Bram. 1897.
\newblock \href {https://www.gutenberg.org/ebooks/345} {\emph{Wuthering
  Heights}}.

\bibitem[{Campos et~al.(2018)Campos, Mangaravite, Pasquali, Jorge, Nunes, and
  Jatowt}]{campos2018yake}
Ricardo Campos, V{\'\i}tor Mangaravite, Arian Pasquali, Al{\'\i}pio~M{\'a}rio
  Jorge, C{\'e}lia Nunes, and Adam Jatowt. 2018.
\newblock Yake! collection-independent automatic keyword extractor.
\newblock In \emph{European Conference on Information Retrieval}, pages
  806--810. Springer.

\bibitem[{Chiang et~al.(2023)Chiang, Li, Lin, Sheng, Wu, Zhang, Zheng, Zhuang,
  Zhuang, Gonzalez, Stoica, and Xing}]{vicuna2023}
Wei-Lin Chiang, Zhuohan Li, Zi~Lin, Ying Sheng, Zhanghao Wu, Hao Zhang, Lianmin
  Zheng, Siyuan Zhuang, Yonghao Zhuang, Joseph~E. Gonzalez, Ion Stoica, and
  Eric~P. Xing. 2023.
\newblock \href {https://vicuna.lmsys.org} {Vicuna: An open-source chatbot
  impressing gpt-4 with 90\%* chatgpt quality}.

\bibitem[{Cox et~al.(1997)Cox, Kilian, Leighton, and Shamoon}]{cox1997secure}
Ingemar~J Cox, Joe Kilian, F~Thomson Leighton, and Talal Shamoon. 1997.
\newblock Secure spread spectrum watermarking for multimedia.
\newblock \emph{IEEE transactions on image processing}, 6(12):1673--1687.

\bibitem[{De~Vleeschouwer et~al.(2002)De~Vleeschouwer, Delaigle, and
  Macq}]{de2002invisibility}
Christophe De~Vleeschouwer, J-F Delaigle, and Benoit Macq. 2002.
\newblock Invisibility and application functionalities in perceptual
  watermarking an overview.
\newblock \emph{Proceedings of the IEEE}, 90(1):64--77.

\bibitem[{Emily(1847)}]{wh}
Brontë Emily. 1847.
\newblock \href {https://www.gutenberg.org/ebooks/768} {\emph{Wuthering
  Heights}}.

\bibitem[{Feng et~al.(2018)Feng, Wallace, Alvin~Grissom, Rodriguez, Iyyer, and
  Boyd-Graber}]{fengpathologies}
Shi Feng, Eric Wallace, II~Alvin~Grissom, Pedro Rodriguez, Mohit Iyyer, and
  Jordan Boyd-Graber. 2018.
\newblock Pathologies of neural models make interpretation difficult.
\newblock In \emph{Empirical Methods in Natural Language Processing}.

\bibitem[{Garg and Ramakrishnan(2020)}]{garg2020bae}
Siddhant Garg and Goutham Ramakrishnan. 2020.
\newblock Bae: Bert-based adversarial examples for text classification.
\newblock In \emph{Proceedings of the 2020 Conference on Empirical Methods in
  Natural Language Processing (EMNLP)}, pages 6174--6181.

\bibitem[{Goldstein et~al.(2023)Goldstein, Sastry, Musser, DiResta, Gentzel,
  and Sedova}]{goldstein2023generative}
Josh~A Goldstein, Girish Sastry, Micah Musser, Renee DiResta, Matthew Gentzel,
  and Katerina Sedova. 2023.
\newblock Generative language models and automated influence operations:
  Emerging threats and potential mitigations.
\newblock \emph{arXiv preprint arXiv:2301.04246}.

\bibitem[{HanSol(2022)}]{webnovel1}
Park HanSol. 2022.
\newblock \href {https://www.koreatimes.co.kr/www/art/2022/04/398_326740.html}
  {Web-based novels ride tide of popularity as sources for webtoon, drama
  adaptations}.
\newblock \emph{The Korea Times}.

\bibitem[{Hao et~al.(2018)Hao, Xiang, Li, Yang, and Shen}]{Hao2018}
Wei Hao, Lingyun Xiang, Yan Li, Peng Yang, and Xiaobo Shen. 2018.
\newblock \href {https://doi.org/10.3970/cmc.2018.03510} {Reversible natural
  language watermarking using synonym substitution and arithmetic coding}.

\bibitem[{He et~al.(2022{\natexlab{a}})He, Xu, Lyu, Wu, and
  Wang}]{he2022protecting}
Xuanli He, Qiongkai Xu, Lingjuan Lyu, Fangzhao Wu, and Chenguang Wang.
  2022{\natexlab{a}}.
\newblock Protecting intellectual property of language generation apis with
  lexical watermark.
\newblock In \emph{Proceedings of the AAAI Conference on Artificial
  Intelligence}, volume~36, pages 10758--10766.

\bibitem[{He et~al.(2022{\natexlab{b}})He, Xu, Zeng, Lyu, Wu, Li, and
  Jia}]{hecater}
Xuanli He, Qiongkai Xu, Yi~Zeng, Lingjuan Lyu, Fangzhao Wu, Jiwei Li, and Ruoxi
  Jia. 2022{\natexlab{b}}.
\newblock Cater: Intellectual property protection on text generation apis via
  conditional watermarks.
\newblock In \emph{Advances in Neural Information Processing Systems}.

\bibitem[{Honnibal and Montani(2017)}]{spacy2}
Matthew Honnibal and Ines Montani. 2017.
\newblock {spaCy 2}: Natural language understanding with {B}loom embeddings,
  convolutional neural networks and incremental parsing.
\newblock To appear.

\bibitem[{Hsu and Wu(1999)}]{hsu1999hidden}
Chiou-Ting Hsu and Ja-Ling Wu. 1999.
\newblock Hidden digital watermarks in images.
\newblock \emph{IEEE Transactions on image processing}, 8(1):58--68.

\bibitem[{Jin et~al.(2020)Jin, Jin, Zhou, and Szolovits}]{jin2020bert}
Di~Jin, Zhijing Jin, Joey~Tianyi Zhou, and Peter Szolovits. 2020.
\newblock Is bert really robust? a strong baseline for natural language attack
  on text classification and entailment.
\newblock In \emph{Proceedings of the AAAI conference on artificial
  intelligence}, volume~34, pages 8018--8025.

\bibitem[{Kingma and Welling(2014)}]{kingmaauto}
Diederik~P Kingma and Max Welling. 2014.
\newblock Auto-encoding variational bayes.
\newblock In \emph{Int. Conf. on Learning Representations}.

\bibitem[{Kirchenbauer et~al.(2023)Kirchenbauer, Geiping, Wen, Katz, Miers, and
  Goldstein}]{kirchenbauer2023watermark}
John Kirchenbauer, Jonas Geiping, Yuxin Wen, Jonathan Katz, Ian Miers, and Tom
  Goldstein. 2023.
\newblock A watermark for large language models.
\newblock \emph{arXiv preprint arXiv:2301.10226}.

\bibitem[{Li et~al.(2021)Li, Zhang, Peng, Chen, Brockett, Sun, and
  Dolan}]{li2021contextualized}
Dianqi Li, Yizhe Zhang, Hao Peng, Liqun Chen, Chris Brockett, Ming-Ting Sun,
  and William~B Dolan. 2021.
\newblock Contextualized perturbation for textual adversarial attack.
\newblock In \emph{Proceedings of the 2021 Conference of the North American
  Chapter of the Association for Computational Linguistics: Human Language
  Technologies}, pages 5053--5069.

\bibitem[{Luo et~al.(2020)Luo, Zhan, Chang, Yang, and
  Milanfar}]{luo2020distortion}
Xiyang Luo, Ruohan Zhan, Huiwen Chang, Feng Yang, and Peyman Milanfar. 2020.
\newblock Distortion agnostic deep watermarking.
\newblock In \emph{Proceedings of the IEEE/CVF Conference on Computer Vision
  and Pattern Recognition}, pages 13548--13557.

\bibitem[{Maas et~al.(2011)Maas, Daly, Pham, Huang, Ng, and
  Potts}]{maas-EtAl:2011:ACL-HLT2011}
Andrew~L. Maas, Raymond~E. Daly, Peter~T. Pham, Dan Huang, Andrew~Y. Ng, and
  Christopher Potts. 2011.
\newblock \href {http://www.aclweb.org/anthology/P11-1015} {Learning word
  vectors for sentiment analysis}.
\newblock In \emph{Proceedings of the 49th Annual Meeting of the Association
  for Computational Linguistics: Human Language Technologies}, pages 142--150,
  Portland, Oregon, USA. Association for Computational Linguistics.

\bibitem[{Merity et~al.(2016)Merity, Xiong, Bradbury, and
  Socher}]{merity2016pointer}
Stephen Merity, Caiming Xiong, James Bradbury, and Richard Socher. 2016.
\newblock Pointer sentinel mixture models.
\newblock \emph{arXiv preprint arXiv:1609.07843}.

\bibitem[{Morris et~al.(2020{\natexlab{a}})Morris, Lifland, Lanchantin, Ji, and
  Qi}]{morris2020reevaluating}
John Morris, Eli Lifland, Jack Lanchantin, Yangfeng Ji, and Yanjun Qi.
  2020{\natexlab{a}}.
\newblock Reevaluating adversarial examples in natural language.
\newblock In \emph{Proceedings of the 2020 Conference on Empirical Methods in
  Natural Language Processing: Findings}, pages 3829--3839.

\bibitem[{Morris et~al.(2020{\natexlab{b}})Morris, Lifland, Yoo, and
  Qi}]{morris2020textattack}
John~X Morris, Eli Lifland, Jin~Yong Yoo, and Yanjun Qi. 2020{\natexlab{b}}.
\newblock Textattack: A framework for adversarial attacks in natural language
  processing.
\newblock \emph{Proceedings of the 2020 EMNLP, Arvix}.

\bibitem[{Mun et~al.(2019)Mun, Nam, Jang, Kim, and Lee}]{mun2019finding}
Seung-Min Mun, Seung-Hun Nam, Haneol Jang, Dongkyu Kim, and Heung-Kyu Lee.
  2019.
\newblock Finding robust domain from attacks: A learning framework for blind
  watermarking.
\newblock \emph{Neurocomputing}, 337:191--202.

\bibitem[{OpenAI(2022)}]{chatgpt}
OpenAI. 2022.
\newblock \href {https://openai.com/blog/chatgpt} {Introducing chatgpt}.

\bibitem[{Potdar et~al.(2005)Potdar, Han, and Chang}]{potdar2005survey}
Vidyasagar~M Potdar, Song Han, and Elizabeth Chang. 2005.
\newblock A survey of digital image watermarking techniques.
\newblock In \emph{INDIN'05. 2005 3rd IEEE International Conference on
  Industrial Informatics, 2005.}, pages 709--716. IEEE.

\bibitem[{Rizzo et~al.(2019)Rizzo, Bertini, and Montesi}]{rizzo2019fine}
Stefano~Giovanni Rizzo, Flavio Bertini, and Danilo Montesi. 2019.
\newblock Fine-grain watermarking for intellectual property protection.
\newblock \emph{EURASIP Journal on Information Security}, 2019(1):1--20.

\bibitem[{Tao et~al.(2014)Tao, Chongmin, Zain, and Abdalla}]{tao2014robust}
Hai Tao, Li~Chongmin, Jasni~Mohamad Zain, and Ahmed~N Abdalla. 2014.
\newblock Robust image watermarking theories and techniques: A review.
\newblock \emph{Journal of applied research and technology}, 12(1):122--138.

\bibitem[{Taori et~al.(2023)Taori, Gulrajani, Zhang, Dubois, Li, Guestrin,
  Liang, and Hashimoto}]{alpaca}
Rohan Taori, Ishaan Gulrajani, Tianyi Zhang, Yann Dubois, Xuechen Li, Carlos
  Guestrin, Percy Liang, and Tatsunori~B. Hashimoto. 2023.
\newblock Stanford alpaca: An instruction-following llama model.
\newblock \url{https://github.com/tatsu-lab/stanford_alpaca}.

\bibitem[{Tina~Fang et~al.(2017)Tina~Fang, Jaggi, and
  Argyraki}]{tina2017generating}
Tina Tina~Fang, Martin Jaggi, and Katerina Argyraki. 2017.
\newblock Generating steganographic text with lstms.
\newblock In \emph{Proceedings of the 55th Annual Meeting of the Association
  for Computational Linguistics-Student Research Workshop}, CONF, pages
  100--106.

\bibitem[{Topkara et~al.(2006{\natexlab{a}})Topkara, Riccardi, Hakkani-T{\"u}r,
  and Atallah}]{topkara2006natural}
Mercan Topkara, Giuseppe Riccardi, Dilek Hakkani-T{\"u}r, and Mikhail~J
  Atallah. 2006{\natexlab{a}}.
\newblock Natural language watermarking: Challenges in building a practical
  system.
\newblock In \emph{Security, Steganography, and Watermarking of Multimedia
  Contents VIII}, volume 6072, pages 106--117. SPIE.

\bibitem[{Topkara et~al.(2005)Topkara, Taskiran, and
  Delp~III}]{topkara2005natural}
Mercan Topkara, Cuneyt~M Taskiran, and Edward~J Delp~III. 2005.
\newblock Natural language watermarking.
\newblock In \emph{Security, Steganography, and Watermarking of Multimedia
  Contents VII}, volume 5681, pages 441--452. SPIE.

\bibitem[{Topkara et~al.(2006{\natexlab{b}})Topkara, Topkara, and
  Atallah}]{topkara2006hiding}
Umut Topkara, Mercan Topkara, and Mikhail~J Atallah. 2006{\natexlab{b}}.
\newblock The hiding virtues of ambiguity: quantifiably resilient watermarking
  of natural language text through synonym substitutions.
\newblock In \emph{Proceedings of the 8th workshop on Multimedia and security},
  pages 164--174.

\bibitem[{Ueoka et~al.(2021)Ueoka, Murawaki, and
  Kurohashi}]{ueoka2021frustratingly}
Honai Ueoka, Yugo Murawaki, and Sadao Kurohashi. 2021.
\newblock Frustratingly easy edit-based linguistic steganography with a masked
  language model.
\newblock In \emph{Proceedings of the 2021 Conference of the North American
  Chapter of the Association for Computational Linguistics: Human Language
  Technologies}, pages 5486--5492.

\bibitem[{Wang et~al.(2001)Wang, Lin, and Lin}]{wang2001image}
Ran-Zan Wang, Chi-Fang Lin, and Ja-Chen Lin. 2001.
\newblock Image hiding by optimal lsb substitution and genetic algorithm.
\newblock \emph{Pattern recognition}, 34(3):671--683.

\bibitem[{Wolfgang et~al.(1999)Wolfgang, Podilchuk, and
  Delp}]{wolfgang1999perceptual}
Raymond~B Wolfgang, Christine~I Podilchuk, and Edward~J Delp. 1999.
\newblock Perceptual watermarks for digital images and video.
\newblock \emph{Proceedings of the IEEE}, 87(7):1108--1126.

\bibitem[{Yang et~al.(2022)Yang, Zhang, Chen, Zhang, Ma, Wang, and
  Yu}]{yang2022tracing}
Xi~Yang, Jie Zhang, Kejiang Chen, Weiming Zhang, Zehua Ma, Feng Wang, and
  Nenghai Yu. 2022.
\newblock Tracing text provenance via context-aware lexical substitution.
\newblock In \emph{Proceedings of the AAAI Conference on Artificial
  Intelligence}, volume~36, pages 11613--11621.

\bibitem[{Yang et~al.(2018)Yang, Guo, Chen, Huang, and Zhang}]{yang2018rnn}
Zhong-Liang Yang, Xiao-Qing Guo, Zi-Ming Chen, Yong-Feng Huang, and Yu-Jin
  Zhang. 2018.
\newblock Rnn-stega: Linguistic steganography based on recurrent neural
  networks.
\newblock \emph{IEEE Transactions on Information Forensics and Security},
  14(5):1280--1295.

\bibitem[{Yang et~al.(2020)Yang, Zhang, Hu, Hu, and Huang}]{yang2020vae}
Zhong-Liang Yang, Si-Yu Zhang, Yu-Ting Hu, Zhi-Wen Hu, and Yong-Feng Huang.
  2020.
\newblock Vae-stega: linguistic steganography based on variational
  auto-encoder.
\newblock \emph{IEEE Transactions on Information Forensics and Security},
  16:880--895.

\bibitem[{Zeyi(2021)}]{webnovel2}
Yang Zeyi. 2021.
\newblock \href {https://www.protocol.com/china/chinese-web-novels-china}
  {China is reinventing the way the world reads}.
\newblock \emph{Protocol}.

\bibitem[{Zhong et~al.(2020)Zhong, Huang, Mastorakis, and
  Shih}]{zhong2020automated}
Xin Zhong, Pei-Chi Huang, Spyridon Mastorakis, and Frank~Y Shih. 2020.
\newblock An automated and robust image watermarking scheme based on deep
  neural networks.
\newblock \emph{IEEE Transactions on Multimedia}, 23:1951--1961.

\bibitem[{Zhu et~al.(2018)Zhu, Kaplan, Johnson, and Fei-Fei}]{zhu2018hidden}
Jiren Zhu, Russell Kaplan, Justin Johnson, and Li~Fei-Fei. 2018.
\newblock Hidden: Hiding data with deep networks.
\newblock In \emph{Proceedings of the European conference on computer vision
  (ECCV)}, pages 657--672.

\bibitem[{Ziegler et~al.(2019)Ziegler, Deng, and Rush}]{ziegler2019neural}
Zachary Ziegler, Yuntian Deng, and Alexander~M Rush. 2019.
\newblock Neural linguistic steganography.
\newblock In \emph{Proceedings of the 2019 Conference on Empirical Methods in
  Natural Language Processing and the 9th International Joint Conference on
  Natural Language Processing (EMNLP-IJCNLP)}, pages 1210--1215.

\end{thebibliography}
\bibliographystyle{acl_natbib}

\newpage

\begin{table}[t]
\begin{adjustbox}{width=\columnwidth}
\begin{tabular}{ccccc}
    \toprule
    Robustness & \makecell{\small{Corr.}\\ \small{Types}} & \makecell{ContextLS\\\citep{yang2022tracing}} & Keyword & Syntactic  \\
    \hline
    \multirow{3}{*}{$\mathcal{R}_{g_1}$} 
                            & D & 0.656 & 0.944 & 0.921   \\
                            & I & 0.608 & 0.955 & 0.959   \\
                            & S & 0.646 & 0.974 & 0.949    \\
    \hline 
    \multirow{3}{*}{$\mathcal{R}_{g_2}$} 
                            & D & 0.649 & 0.679 & 0.535    \\
                            & I & 0.591 & 0.679 & 0.517     \\
                            & S & 0.641 & 0.756 & 0.612      \\
    \bottomrule
\end{tabular}
\label{tab:final_bd}
\end{adjustbox}
\caption{Robustness of $g_1$ and $g_2$ for three components against three corruption types: Deletion (D), Insertion (I), and Substitution (S) under 5\% corruption rate on IMDB.}
\label{table:robustness_full}
\end{table}

\begin{table}[t]
\begin{adjustbox}{width=\columnwidth}
\begin{tabular}{cccc}
    \toprule
    $\mathcal{R}_{g_1}$ & \makecell{\small{Corr.}\\ \small{Types}} &  Keyword & Syntactic  \\
    \hline
    \multirow{3}{*}{Wikitext-2} 
                            & D & 0.878 & 0.871   \\
                            & I & 0.909 & 0.939   \\
                            & S & 0.935 & 0.963    \\
    \hline
    \multirow{3}{*}{Dracula} 
                            & D & 0.947 & 0.940   \\
                            & I & 0.953 & 0.972   \\
                            & S & 0.987 & 0.963    \\
    \hline
    \multirow{3}{*}{WH} 
                            & D & 0.945 & 0.934   \\
                            & I & 0.963 & 0.965   \\
                            & S & 0.977 & 0.936    \\

    \bottomrule
\end{tabular}
\end{adjustbox}
\caption{Robustness of $g_1$ on our proposed components against three corruption types: Deletion (D), Insertion (I), and Substitution (S) under 5\% corruption rate.}
\label{table:robustness1_other_datasets}
\end{table}

\begin{table}[t]
\centering
\begin{adjustbox}{width=0.8\columnwidth}
\begin{tabular}{cc|cc}
    \toprule
     \multicolumn{2}{c}{Metrics} & With CC & Discarding CC  \\
    \hline 
   \multicolumn{4}{c}{\textbf{IMDB}} \\ 
    \hline \hline 
    BPW ($\uparrow$) & & 0.130 & 0.151  \\
    \hline
    \multirow{3}{*}{\makecell{BER($\downarrow$)\\@CR=0.025}} 
                & D  & 0.072 & 0.085 \\
                & I  & 0.113 & 0.123 \\
                & S  & 0.111 & 0.125 \\
    \hline
    \multirow{3}{*}{\makecell{BER($\downarrow$)\\@CR=0.05}} 
            & D  & 0.195 & 0.224 \\
            & I  & 0.161 & 0.194 \\
            & S  & 0.187 & 0.200 \\
    \hline
    ES ($\uparrow$)&& 0.970  & 0.963 \\
    SS ($\uparrow$)&& 0.974  & 0.978 \\
    \hline
    \multicolumn{4}{c}{\textbf{Wikitext-2}} \\ 
    \hline \hline 
    BPW ($\uparrow$) & & 0.099 & 0.115  \\
    \hline
    \multirow{3}{*}{\makecell{BER($\downarrow$)\\@CR=0.025}} 
                & D  & 0.137 & 0.132 \\
                & I  & 0.197 & 0.180 \\
                & S  & 0.142 & 0.140 \\
    \hline
    \multirow{3}{*}{\makecell{BER($\downarrow$)\\@CR=0.05}} 
            & D  & 0.274 & 0.231 \\
            & I  & 0.195 & 0.172 \\
            & S  & 0.194 & 0.179 \\
    \hline
    ES ($\uparrow$)&& 0.966  & 0.961 \\
    SS ($\uparrow$)&& 0.993  & 0.993 \\
    \hline
        \multicolumn{4}{c}{\textbf{Dracula}} \\ 
    \hline \hline 
    BPW ($\uparrow$) & & 0.146 & 0.135  \\
    \hline
    \multirow{3}{*}{\makecell{BER($\downarrow$)\\@CR=0.025}} 
                & D  & 0.030 & 0.062 \\
                & I  & 0.063 & 0.093 \\
                & S  & 0.081 & 0.099 \\
    \hline
    \multirow{3}{*}{\makecell{BER($\downarrow$)\\@CR=0.05}} 
            & D  & 0.177 & 0.193 \\
            & I  & 0.155 & 0.234 \\
            & S  & 0.164 & 0.179 \\
    \hline
    ES ($\uparrow$)&& 0.963  & 0.944 \\
    SS ($\uparrow$)&& 0.971  & 0.965 \\
    \hline
            \multicolumn{4}{c}{\textbf{Wuthering Heights}} \\ 
    \hline \hline 
    BPW ($\uparrow$) & & 0.114 & 0.113  \\
    \hline
    \multirow{3}{*}{\makecell{BER($\downarrow$)\\@CR=0.025}} 
                & D  & 0.063 & 0.075 \\
                & I  & 0.068 & 0.114 \\
                & S  & 0.096 & 0.117 \\
    \hline
    \multirow{3}{*}{\makecell{BER($\downarrow$)\\@CR=0.05}} 
            & D  & 0.169 & 0.204 \\
            & I  & 0.133 & 0.200 \\
            & S  & 0.161 & 0.190 \\
    \hline
    ES ($\uparrow$)&& 0.964  & 0.942 \\
    SS ($\uparrow$)&& 0.975  & 0.969 \\
    \hline
\end{tabular}
\end{adjustbox}
\caption{Watermarking embedding and extraction results after discarding the coordination dependency on IMDB.}
\label{tab:dep-wo-cc-full}
\end{table}

\begin{table}[b]
\begin{adjustbox}{width=\columnwidth}
\begin{tabular}{cccc}
    \toprule
             & \small Hyperparm. &  Keyword & Syntactic  \\
    \hline
    \multirow{2}{*}{IMDB} 
                            & KR & 0.06 & 0.05   \\
                            & $k_2$ & 4 & 4   \\
    \multirow{2}{*}{Wikitext-2} 
                            & KR & 0.06 & 0.07   \\
                            & $k_2$ & 4 & 4*   \\
    \multirow{2}{*}{Dracula} 
                            & KR & 0.07 & 0.03   \\
                            & $k_2$ & 4 & 3   \\
    \multirow{2}{*}{WH} 
                            & KR & 0.05 & 0.03   \\
                            & $k_2$ & 4 & 4   \\
    \hline

    \bottomrule
\end{tabular}
\end{adjustbox}
\caption{Configurations used in each dataset to ensure payload around BPW=0.1. KR denotes the ratio of keyword to the number of words in the sentence. We ensure at least one keyword is selected in each sentence.}
\label{tab:configuration}
\end{table}

\begin{figure}[t]
    \centering
    \includegraphics[width=0.45\textwidth]{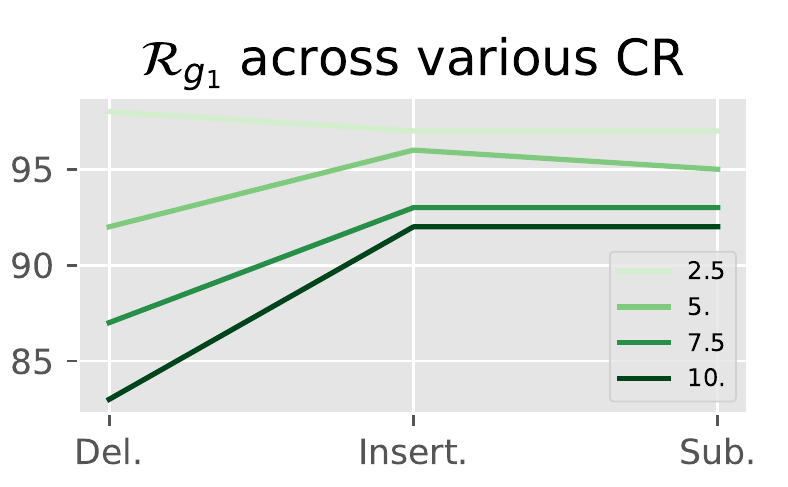}
    \caption{Robustness of $g_1$ at higher corruption rate.} 
    \label{fig:high-cr}
\end{figure}

\appendix
\section{Appendix}
\label{sec:appendix}

\subsection{Implementation Details}
\noindent\textbf{Dataset Split}
Following ContextLS, we subsampled the first 5000 sentences and used the same subset across all methods. Our preliminary experiments showed subsampling other samples only led to minor variability: standard error of the mean BPW across 3 trials 0.002. We use the same subset for all our experiments to avoid any confounding factors. For the robustness experiment, which had a stochastic element, the standard errors for BER’s for insertion and substitution were also marginal (both 0.004) compared to the performance gap. 

To finetune our robust infill model, we required a train set other than the test set that will be watermarked. For IMDB and Wikitext-2, we used the original training split. For the novels datasets, we take the first 40\% of the text as the train set and the rest as the test set. The same splits are also used for training AWT as well.

\noindent\textbf{Corruption}
To test the robustness, we corrupt the first 1000 sentences of the 5000 test sets. Since the watermark embedding processes for ours and ContextLS are deterministic given the message, we run the embedding experiment once for a fixed random seed. Due to the implementation of TextAttack, some corruption modules may be non-deterministic, which will lead to a non-deterministic BER. We find that the deletion module we used is deterministic so we run the robustness experiment once. On the other hand, we create five corrupted samples per sample for insertion and substitution and report the mean for ours and ContextLS.

\noindent\textbf{Computation Time}
The actual watermarking process does not require gradient computation. The largest bottleneck in the pipeline is the forward passes of the infill model. The actual wall clock time and the number of passes are detailed on \cref{subsec:discussion_with_cls}. Training the infill model requires the most computation time. We finetune all our models in a single GPU environment using either Titan RTX or RTX 3090. Finetuning on Wikitext-2 was the longest among the datasets, which required approximately 22 GPU-hours for 100 epochs. 

\noindent\textbf{Training Details of Infill Model}
We use AdamW with a learning rate of 5e-5 using linear warmup 0.1 of the total training steps. All our models are trained for 100 epochs and we used the last checkpoint. For random masking, we simply mask out 15\% of the words using whole word masking strategy.

\subsection{AWT Implementation Details}
\label{appendix:awt}
We use the official implementation and mostly adhere to the hyperparameters employed by AWT unless otherwise noted.
In the original paper, the experiment was conducted only for a lower payload BPW=0.05 on the Wikitext-2 dataset, so implementation details for a higher payload BPW=0.1 or other datasets needed to be adjusted.

First, we replaced the AWD-LSTM language model with GPT-2, providing a superior language modeling capability. Second, when the payload was increased to BPW=0.1, the weighting term for the reconstruction loss (see Section IV-D)  was doubled at the second training stage of AWT to make the model converge. 
Third, we combined data for Dracula and Wuthering Heights into a single dataset to train and evaluate the AWT model because we were unable to train the model for each dataset separately due to a lack of data.

For a fair comparison in robustness experiments, watermarked segments are concatenated and then split into sentences, to which corruption is applied on a per-sentence basis. Lastly, the corrupted segments are used to report BER against attacks.
In addition, AWT constructs a dictionary of tokens using the corpus before watermarking embedding. This may introduce unknown tokens for insertion and substitution, in which case we exclude these tokens.

\subsection{AWT on IMDB dataset}
\label{appendix:awt-imdb}
The text reconstruction loss did not converge for the IMDB datasets. This led to a severe quality decrease in the watermarked sentence as shown below in Table \ref{tab:awt-imdb}. We nevertheless test the robustness under corruption. The BER@CR=0.05 for the three corruption types were 0.283, 0.278, and 0.299.

\begin{table}[h]
\begin{tabular}{p{0.97\linewidth}}
    \hrule 
    \centering \textbf{Original and  Watermarked} \tabularnewline
    \small{"Budget limitations, time restrictions, shooting a script and then cutting it, cutting it, cutting it... This crew is a group of good, young filmmakers; }\\[5pt]
   \small political/strategic Show time *very shooting a script and then cutting it, cutting it, cutting it... This crew is a group of good, young Gilbert 
   \hrule
\end{tabular}
\caption{Example of failing to reconstruct the cover text for AWT on IMDB.}
\vspace{-3mm}
\label{tab:awt-imdb}
\end{table}

\subsection{More Results}
\label{appendix:nli-ordering}
\textbf{Ordering of NLI and Discarding Coordination}
To define the ordering of syntactic dependency, we mask out each of the dependencies on the train set and then infill the masked-out dependencies. The infilled sentences are compared with the original sentence. A Pythonic algorithm for one sample is shown Alg. \ref{alg1}. This is done for 500 samples of IMDB. The resultant ordering is shown in Table \ref{tab:dep-list}.

As discussed in \cref{subsec:pitfall}, substituting the coordination dependency (CC) is often leads to a semantic drift that is undetectable by automatic metrics. We also provide the BPW and robustenss results after discarding CC from the NLI ordering list in Table \ref{tab:dep-wo-cc-full}.

\noindent\textbf{Character-based Corruption}
We also experiment with character-based corruption, which may happen when unintentionally during manual transcription. We simulate this type of corruption by randomly swapping a character with a neighboring character using TextAttack. Similar to our main experiment, we test on CR=\{2.5\%, 5\%\}. On the IMDB dataset, our Syntactic Dependency Component model has a BER of .079 and .167, respectively. While our RI model did not explicitly train on this type of error, it nevertheless improves robustness to  0.063 and 0.142, respectively.

\noindent\textbf{ContextLS + Robust Infill}
Using a finetuned infill model gave a meaningful boost in robustness in all datasets for our method. Is this model effective for ContextLS as well? Using an infill model trained using random masks is not always beneficial to the robustness of ContextLS and the improvement is marginal compared to that of ours (Appendix Table \ref{tab:CLS_comparison}). This is expected given our analysis in \cref{subsec:Phase1} that Phase 1 is a strong bottleneck for ContextLS, yet we believe it can be further improved if a specific masking strategy used in ContextLS is adapted when finetuning the infill model.

\subsection{More Discussions}
\label{appendix:more_discussion}

\noindent\textbf{Computing BER}
For ours and ContextLS, the number of bits varies by sentence. This leads to an issue when computing BER as the predicted message may have less or more bits than the true message. To accurately assess BER, we assume that the true number of bits is unknown during extraction. When the extracted number of bits is less than the ground truth, we consider all unpredicted bits as errors. Conversely, when more bits are extracted, we truncate them and consider all over-extracted bits as errors.    

\subsection{Human Evaluation}
\label{appendix:human-eval}
\begin{figure*}[h]
    \centering
    \includegraphics[width=0.9\textwidth,height=\textheight,keepaspectratio]{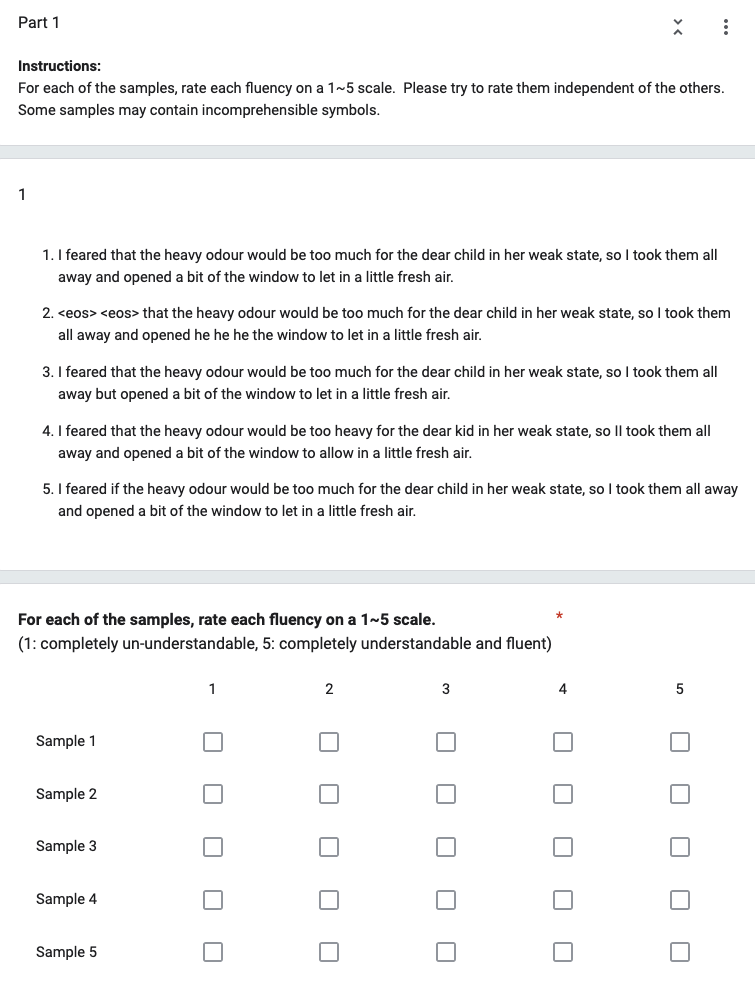}
    \caption{A screenshot of human evaluation survey evaluating fluency.} 
    \label{fig:survey}
    \vspace{-5mm}
\end{figure*}

We collected human annotations of the watermarked texts through ClickWorker and disclosed the responses may be used for research purposes. The workers were recruited from United States, United Kingdom, and Ireland at the age of 20-99 who considered themselves with English as their native languages. The survey was designed to take approximately 40-60 minutes and the fee was 20 Euros, which was over the minimum wages of the three countries. We only used the responses that had an adequately high "semantic was completely maintained" answer proportion for those watermarked texts that were not altered from the cover text to ensure the instructions were followed. When thresholding this proportion by 0.5, 2 responses were discarded out of the 7 responses. Screenshots of the survey are in the last page in Figure \ref{fig:survey}. 
The survey consisted of 10 random samples each from Dracula and Wuthering Heights. We excluded Wikitext-2 as AWT preprocessed the name of the entities as unknown tokens, which may lead to substantial decrease in fluency for the annotators. IMDB was excluded as the text reconstruction loss did not converge for AWT, which led to incomprehensible sentences. Part 1 consisted of rating the fluency of each sentence including the original cover text. Fluency $\Delta$ was computed by subtracting the fluency of the watermarked sample from the original one. Part 2 consisted of rating how much semantics is maintained given the reference sentence (cover text). 

\begin{table}
\centering \textbf{Types of Dependencies}
\hrule
\begin{multicols}{3}
\begin{enumerate}
    \item expl
    \item cc
    \item auxpass
    \item agent 
    \item mark
    \item aux
    \item prep
    \item det
    \item prt
    \item parataxis
    \item predet
    \item case 
    \item csubj
    \item acl 
    \item advcl     
\end{enumerate}
\end{multicols}
\hrule 
\caption{List of dependencies ordered by NLI entail score (Top-15). For details of each dependency, please refer to the \href{https://downloads.cs.stanford.edu/nlp/software/dependencies_manual.pdf}{ Stanford Dependencies Manual.}}
\label{tab:dep-list}
\end{table}

\begin{table}[t]
\begin{adjustbox}{width=\columnwidth}
\begin{tabular}{lc cccc}
    \toprule
    Dataset & Metric & Keyword & Syntactic & +RI & -NLI Ord.   \\
    \hline
    \multirow{2}{*}{D1}
        & ES & 0.932 & 0.975 &  0.975 & 0.854  \\
        & SS & 0.967 & 0.982 &  0.981 & 0.946 \\ 
    \hline
    \multirow{2}{*}{D2}
        & ES & 0.895 & 0.966 &  0.966 & 0.696 \\
        & SS & 0.979 & 0.993 &  0.993 & 0.953 \\ 
    \hline
    \multirow{2}{*}{D3}
        & ES & 0.920 & 0.960 & 0.963  & 0.835\\
        & SS & 0.964  & 0.974 &  0.971  & 0.939\\ 
    \hline
    \multirow{2}{*}{D4}
        & ES & 0.910 & 0.964 & 0.964 & 0.790 \\
        & SS & 0.967 & 0.976 & 0.975 & 0.941 \\ 
    \bottomrule
\end{tabular}
\end{adjustbox}
\caption{Semantic scores (ES: entailment score, SS: semantic similarity) of the watermarked sets in for variants of our method.} 
\label{tab:semantic_ours}
\end{table}

\begin{table}[h]
\begin{adjustbox}{width=\columnwidth}
\begin{tabular}{cc|cc|cc}
    \toprule
     \multicolumn{2}{c}{Metrics} & ContextLS & $\Delta$ & Ours & $\Delta$ \\
    \hline 
    BPW ($\uparrow$) & & 0.100 & +0.0 & 0.130  &  +1.3\%   \\
    \hline
    \multirow{3}{*}{\makecell{BER($\downarrow$)\\@CR=0.025}} 
                & D  & 0.219 & \textcolor{NavyBlue}{+2.0\% }  &  0.100 & \textcolor{NavyBlue}{+2.8\%}\\
                & I  & 0.303 & -0.5\% & 0.153 & \textcolor{NavyBlue}{+4.0\%}\\
                & S  & 0.273 & \textcolor{NavyBlue}{+1.6\%} & 0.133 & \textcolor{NavyBlue}{+2.2\%} \\
    \hline
        \multirow{3}{*}{\makecell{BER($\downarrow$)\\@CR=0.05}} 
                & D  & 0.392 & \textcolor{NavyBlue}{+1.4\%} & 0.279 & \textcolor{NavyBlue}{+9.4\%} \\
                & I  & 0.362 & \textcolor{NavyBlue}{+2.0\%} & 0.236 & \textcolor{NavyBlue}{+7.9\%} \\
                & S  & 0.343 & 0.0\% & 0.224 & \textcolor{NavyBlue}{+4.5\%} \\
    \hline

\end{tabular}
\end{adjustbox}
\caption{The effect of using Robust Infill (RI) model on ContextLS on the first 1,000 sentences of IMDB. A \textcolor{NavyBlue}{positive number} denotes improvement in BER. For reference, we show the improvement in ours.}
\label{tab:CLS_comparison}
\end{table}

\newpage
\mbox{~}

\subsection{Watermarked Examples}
\label{appendix:examples}
Examples of watermarked texts are provided in Table \ref{tab:example2}-\ref{tab:example1}. The watermarked words are marked by color. 
For ours and ContextLS, some texts may be unaltered from the cover text if the original text is included in the valid watermarked sets. For AWT, this is only possible if the watermark has been embedded at a different section of the segment since it usually takes multiple sentences (40 words) as inputs. Thus, we display only those examples that have been modified for qualitative analysis. (Conversely, for human evaluation, we randomly sample sentences.) For Wikitext-2, which contains considerable amount of entities, many of the entities have been marked as unknown tokens on AWT outputs. We manually substitute these tokens for presentation purposes. 

\begin{table*}[t]
\begin{adjustbox}{height=0.45\textheight, center}
\begin{tabular}{p{0.9\textwidth}}
    \multicolumn{1}{c}{\textbf{Dracula}} \\ 
    \textit{Original} \\
    \textit{Ours} \\
    \textit{Ours (Discarding CC)} \\
    \textit{Context-LS} \\
    \textit{AWT} \\
    \hrule
    I feared that the heavy odour would be too much for the dear child in her weak state, so I took them all away and opened a bit of the window to let in a little fresh air.	\\\\
    I feared that the heavy odour would be too much for the dear child in her weak state, so I took them all away \emp{but} opened a bit of the window to let in a little fresh air.	\\\\
    I feared \emp{if} the heavy odour would be too much for the dear child in her weak state, so I took them all away and opened a bit of the window to let in a little fresh air.	\\\\
    I feared that the heavy odour would be too \emp{heavy} for the dear \emp{kid} in her weak state, so \emp{II} took them all away and opened a bit of the window to \emp{allow} in a little fresh air.	\\\\
    \emp{<eos> <eos>} that the heavy odour would be too much for the dear child in her weak state, so I took them all away and opened \emp{he he he} the window to let in a little fresh air.\\
    \hrule
    In the hall he opened the dining-room door, and we passed in, he closing the door carefully behind him.	\\
    In the hall he opened the dining-room door, \emp{as} we passed in, he closing the door carefully behind him.	\\
    In the hall he opened the dining-room door, and we passed in, he closing the door carefully behind him.	\\
    In the hall he opened the dining-room door, and we passed in, he closing the door carefully behind him.	\\
    In the hall \emp{I} opened the dining-room door, and we passed in, \emp{on} closing the door carefully behind him.
    \hrule
    He had evidently read it, and was thinking it over as he sat with his hand to his brow.	\\
    He had evidently read it, and was thinking it over as he sat with his hand to his brow.	\\
    He had evidently read it, and was thinking it over \emp{while} he sat with his hand to his brow.	\\
    He had evidently read it, and was thinking it over as he sat with his hand to his \emp{head}.	\\
    He had evidently read it, and was thinking it over to he sat with \emp{the} hand to the \emp{Dress}.
    \hrule
    I had done my part, and now my next duty was to keep up my strength.	\\
    I had done my part, \emp{but} now my next duty was to keep up my strength.	\\
    I \emp{was} done my part, and now my next duty was to keep up my strength.	\\
    I had \emp{performed} my part, and now my \emp{new} duty was to keep up my strength.	\\
    I had done my part, and now my next duty was \emp{keep} keep up my strength. \\
    \hrule
    I weren't a-goin' to fight, so I waited for the food, and did with my 'owl as the wolves, and lions, and tigers does.	\\
    I weren't a-goin' to fight, so I waited for the food, \emp{or} did with my 'owl as the wolves, and lions, and tigers does.\\
    I weren't a-goin' to fight, so I waited for the food, and did with my 'owl as the wolves, and lions, and tigers does.	\\
    I weren't a-goin'to fight, so I waited for the food, and did with my ‘owl as the wolves, and lions, and tigers does.	\\
    \emp{<eos>} weren't \emp{chased} to fight, so \emp{<eos>} waited for the food, and did with my 'owl as the wolves, and lions, and tigers does.
    \hrule

\end{tabular}
\end{adjustbox}
\caption{Samples of watermarked texts. The original cover text is shown in the first row.}
\label{tab:example2}
\end{table*}

\begin{table*}[t]
\begin{adjustbox}{height=0.45\textheight, center}
\begin{tabular}{p{0.9\textwidth}}
    \multicolumn{1}{c}{\textbf{Wuthering Heights}} \\ 
    \textit{Original} \\
    \textit{Ours} \\
    \textit{Ours (Discarding CC)} \\
    \textit{Context-LS} \\
    \textit{AWT} \\
    \hrule
    “In general I’ll allow that it would be, Ellen,” she continued; “but what misery laid on Heathcliff could content me, unless I have a hand in it?\\ 
    “In general I’ll allow that it would be, Ellen,” she continued; “\emp{and} what misery laid on Heathcliff could content me, unless I have a hand in it?\\ 
    “In general I’ll allow that it would be, Ellen,” she continued; “but what misery laid on Heathcliff could content me, unless I have a hand in it?\\ 
    “In general I’ll allow that it would be, Ellen,” she continued; “but what misery laid on Heathcliff could content me, unless I have a hand in it?\\ 
    \emp{that} “In general I’ll allow that it would be, Ellen,” she continued; “but what misery laid on Heathcliff could content me, unless I have a hand in it?
    \hrule
    He took her education entirely on himself, and made it an amusement.	\\
    He took her education entirely on himself, \emp{but} made it an amusement.	\\
    He took her education entirely \emp{for} himself, and made it an amusement.	\\
    He took her \emp{schooling} entirely on himself, and made it an amusement.	\\
    He took her education entirely on himself, and made it an amusement.\\
    \hrule
    I’m sure you would have as much pleasure as I in witnessing the conclusion of the fiend’s existence; he’ll be your death unless you overreach him; and he’ll be my ruin.	\\\\
    I’m sure you would have as much pleasure as I in witnessing the conclusion of the fiend’s existence; he’ll be your death unless you overreach him; and he’ll be my ruin.	\\\\
    I’m sure you would have as much pleasure as I in witnessing the conclusion of the fiend’s existence; he’ll be your death \emp{if} you overreach him; and he’ll be my ruin.	\\\\
    I’m sure you would have as much pleasure as \emp{mine} in witnessing the conclusion of the fiend’s \emp{presence}; he’ll be your death unless you overreach him; and he’ll be my ruin.	\\\\
    I’m sure you would have as much pleasure as \emp{as} in witnessing the conclusion \emp{as} the fiend’s existence; \emp{as} be your death unless you overreach him; and he’ll be \emp{polyglot,} ruin.\\
    \hrule
    To my joy, he left us, after giving this judicious counsel, and Hindley stretched himself on the hearthstone.	\\ 
    To my joy, he left us, after giving this judicious counsel, \emp{while} Hindley stretched himself on the hearthstone.	\\ 
    \emp{With} my joy, he left us, after giving this judicious counsel, and Hindley stretched himself on the hearthstone.	\\ 
    To my joy, he left us, after *delivering* this judicious counsel, and Hindley stretched himself on the hearthstone.	\\ 
    To my joy, \emp{over} left us, after giving this judicious counsel, and Hindley stretched himself \emp{<eos>} the hearthstone.
    \hrule
    I heard my master mounting the stairs—the cold sweat ran from my forehead: I was horrified.	\\
    I heard my master mounting the stairs—the cold sweat ran \emp{across} my forehead: I was horrified.\\
    I heard my master mounting the stairs—the cold sweat ran \emp{over} my forehead: I was horrified.\\
    I heard my master mounting the stairs— the cold sweat ran from my forehead: I was horrified.\\
    \emp{of} heard my master mounting the stairs—the cold sweat ran from my forehead: I was horrified.\\
    \hrule
\end{tabular}
\end{adjustbox}
\caption{Samples of watermarked texts. The original cover text is shown in the first row.}
\label{tab:example3}
\end{table*}

\begin{table*}[t]
\begin{adjustbox}{height=0.35\textheight, center}
\begin{tabular}{p{0.9\textwidth}}
    \multicolumn{1}{c}{\textbf{Wikitext-2}} \\ 
    \textit{Original} \\
    \textit{Ours} \\
    \textit{Ours (Discarding CC)} \\
    \textit{Context-LS} \\
    \textit{AWT} \\
    \hrule
    He was relieved by Yan Wu, a friend and former colleague who was appointed governor general at Chengdu.	\\
    He was relieved by Yan Wu, a friend and former colleague who was appointed governor general at Chengdu.	\\
    He was relieved by Yan Wu, a friend and former colleague who was appointed governor general at Chengdu.	\\
    He was relieved by Yan Wu, a friend and \emp{ex} colleague who was \emp{named} governor general at Chengdu.\\
    He was relieved \emp{an} Yan Wu , a friend and former colleague who was appointed governor general at Chengdu. \\     
    \hrule
    Keiser decided that this situation made it advisable to control and direct the divided division as two special forces.	\\
    Keiser decided that this situation made it advisable to control and direct the divided division as two special forces.	\\
    Keiser decided \emp{because} this situation made it advisable to control and direct the divided division as two special forces.	\\
    Keiser decided that this situation made it advisable to control and direct the divided unit as two special forces. \\
    Keiser decided that this situation made it advisable to control and direct the divided \emp{division his} two special forces
    \hrule
    His greatest ambition was to serve his country as a successful civil servant, but he proved unable to make the necessary accommodations.\\\\
    His greatest ambition was to serve his country as a successful civil servant, \emp{although} he proved unable to make the necessary accommodations.\\\\
    His greatest ambition was to serve his country \emp{with} a successful civil servant, but he proved unable to make the necessary accommodations .\\\\
    His greatest ambition was to serve his \emp{nation} as a successful civil servant, but he proved unable to make the necessary accommodations.\\
    His greatest ambition was to serve his country \emp{having} a successful civil servant, but he proved unable to make the necessary accommodations.
    \hrule
\end{tabular}
\end{adjustbox}
\caption{Samples of watermarked texts. The original cover text is shown in the first row.}
\label{tab:example4}
\end{table*}

\begin{table*}[t]
\begin{adjustbox}{height=0.5\textheight, center}
\begin{tabular}{p{0.9\textwidth}}
    \multicolumn{1}{c}{\textbf{IMDB}} \\ 
    \textit{Original} \\
    \textit{Ours} \\
    \textit{Ours (Discarding CC)} \\
    \textit{Context-LS} \\
    \hrule
     Photographer Gary(David Hasselhoff)is taking pictures for Linda(Catherine Hickland whose voice and demeanor resemble EE-YOR of the Winnie the Poo cartoon), a virgin studying witchcraft, on the island resort without permission. \\
    \\
    Photographer Gary(David Hasselhoff)is taking pictures for Linda(Catherine Hickland whose voice \emp{or} demeanor resemble EE-YOR of the Winnie the Poo cartoon), a virgin studying witchcraft, on the island resort without permission. \\
    \\
    Photographer Gary(David Hasselhoff)is taking pictures \emp{with} Linda(Catherine Hickland whose voice and demeanor resemble EE-YOR of the Winnie the Poo cartoon), a virgin studying witchcraft, on the island resort without permission. \\	
    \\
    Photographer Gary(David Hasselhoff) is \emp{shooting} pictures for Linda(Catherine Hickland whose voice and demeanor resemble EE-YOR of the Winnie the Poo cartoon), a virgin studying witchcraft, on the island resort without permission. \\ 
    \hrule
     It is amateur hour on every level. \\
     It is amateur hour \emp{of} every level.	\\
    It is amateur hour \emp{of} every level.	\\
     It is amateur hour on every \emp{floor}. \\
    \hrule
    A film that had a lot of potential that was probably held back by it's budget. \\
     A film that had a lot of potential that was probably held back by it's budget. \\	
     A film that had a lot of potential that \emp{is} probably held back by it's budget.\\	
     A film that had a lot of potential that was probably held back by it's budget.\\
    \hrule 
     A gathering of people at a Massachusetts island resort are besieged by the black magic powers of an evil witch killing each individual using cruel, torturous methods.	 \\\\
     A gathering of people at a Massachusetts island resort \emp{was} besieged by the black magic powers of an evil witch killing each individual using cruel, torturous methods. \\\\ 
     A gathering of people at a Massachusetts island resort \emp{is} besieged by the black magic powers of an evil witch killing each individual using cruel, torturous methods.	\\\\      
     A gathering of people at a Massachusetts island resort are besieged by the black magic powers of an evil witch killing each individual using cruel, torturous methods. \\\\
    \hrule
    I have not seen any other movies from the "Crime Doctor" series, so I can't make any comparisons. \\\\
    I have not seen any other movies from the "Crime Doctor" series, \emp{and} I can't make any comparisons. \\\\	
    I have not seen any other movies from the "Crime Doctor" series, so I can't make any comparisons. \\\\	
    I have not seen any other movies from the "\emp{Criminal} Doctor" series, so I can't make any comparisons. \\\\
    \hrule
\end{tabular}
\end{adjustbox}
\caption{Samples of watermarked texts. The original cover text is shown in the first row.}
\label{tab:example1}
\end{table*}

\begin{figure*}[h]
    \centering
    \includegraphics[width=0.9\textwidth,height=\textheight,keepaspectratio]{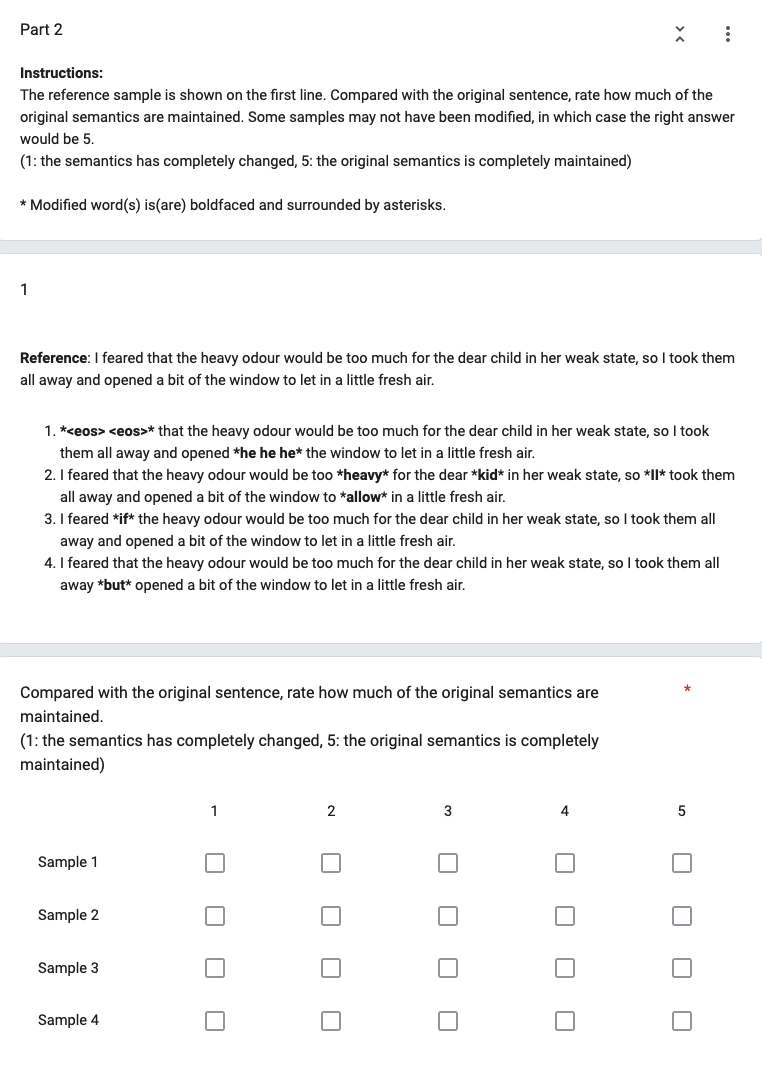}
    \caption{A screenshot of human evaluation survey evaluating semantics compared to the original cover text.} 
    \label{fig:survey2}
    \vspace{-5mm}
\end{figure*}

\end{document}